\documentclass[journal]{IEEEtran}
\usepackage{color}

\usepackage{cite}

\ifCLASSINFOpdf
\usepackage[pdftex]{graphicx}
\graphicspath{{figure/}}
\else
\usepackage{graphicx}
\graphicspath{{figure/}}
\fi
\usepackage{algorithmic}
\usepackage{algorithm}

\newcommand{\ie}{\textit{i}.\textit{e}.}
\newcommand{\eg}{\textit{e}.\textit{g}.}

\usepackage[caption=false,font=footnotesize]{subfig}

\usepackage{multirow}

\begin{document}
%
\title{Superpixel Segmentation Based on Spatially Constrained Subspace Clustering}
%
%
%

\author{Hua Li,
        Yuheng Jia,
        Runmin Cong,~\IEEEmembership{Member,~IEEE,}
        Wenhui Wu,
        Sam Kwong,~\IEEEmembership{Fellow,~IEEE,}
        and Chuanbo Chen

\thanks{This work was supported by the Key Project of Science and Technology Innovation 2030 supported by the Ministry of Science and Technology of China under Grant 2018AAA0101301, in part by the Natural Science Foundation of China under Grants 61772344, 62002014, in part by the Hong Kong RGC General Research Funds under 9042816 (CityU 11209819), in part by the Beijing Nova Program under Grant Z201100006820016, in part by the Fundamental Research Funds for the Central Universities under Grant 2019RC039, in part by Elite Scientist Sponsorship Program by the Beijing Association for Science and Technology, in part by Hong Kong Scholars Program, and in part by China Postdoctoral Science Foundation under Grant 2020T130050, Grant 2019M660438. (Corresponding author: Sam Kwong.)}

\thanks{H. Li is with the School of Software Engineering, Huazhong University of Science and Technology, Wuhan 430074, China, and also with the Department of Computer Science, City University of Hong Kong, Kowloon, Hong Kong (e-mail: huali27-c@my.cityu.edu.hk).}

\thanks{Y. Jia is with the School of Computer Science and Engineering, Southeast University, Nanjing, 211189, China (e-mail: yhjia@seu.edu.cn).}

\thanks{R. Cong is the Institute of Information Science, Beijing Jiaotong University, Beijing 100044, China, and also with the Beijing Key Laboratory of Advanced Information Science and Network Technology, Beijing Jiaotong University, Beijing 100044, China (rmcong@bjtu.edu.cn).}

\thanks{W. Wu is with the College of Electronics and Information Engineering, Shenzhen University, Shenzhen 518060, China (e-mail: wuwenhui@szu.edu.cn).}

\thanks{S. Kwong is with the Department of Computer Science, City University of Hong Kong, Kowloon, Hong Kong, and also with the City University of Hong Kong Shenzhen Research Institute, Shenzhen 518057, China (e-mail: cssamk@cityu.edu.hk).}

\thanks{C. Chen is with the School of Software Engineering, Huazhong University of Science and Technology, Wuhan 430074, China (e-mail: chuanboc@163.com).}
}

\maketitle

\begin{abstract}
Superpixel segmentation aims at dividing the input image into some representative regions containing pixels with similar
 and consistent intrinsic properties, without any prior knowledge about the shape and size of each superpixel.
  In this paper, to alleviate the limitation of superpixel segmentation applied in practical industrial tasks that
  detailed boundaries are difficult to be kept, we regard each representative region with independent semantic
   information as a subspace, and correspondingly formulate superpixel segmentation as a subspace clustering
    problem to preserve more detailed content boundaries.
    We show that a simple integration of superpixel segmentation with the conventional subspace clustering
     does not effectively work due to the spatial correlation of the pixels within a superpixel,
      which may lead to boundary confusion and segmentation error when the correlation is ignored.
      Consequently, we devise a spatial regularization and propose a novel convex locality-constrained
      subspace clustering model that is able to constrain the spatial adjacent pixels with similar attributes
      to be clustered into a superpixel and generate the content-aware superpixels with more detailed boundaries.
      Finally, the proposed model is solved by an efficient alternating direction method of multipliers (ADMM) solver.
      Experiments on different standard datasets demonstrate that the proposed method achieves superior performance both quantitatively and qualitatively compared with some state-of-the-art methods.

\end{abstract}


%
\IEEEpeerreviewmaketitle

\section{Introduction}
%
%
%
%
Superpixels are alternative primitives to represent the original pixels in an image based on certain intrinsic properties, such as coherent spatial relationship, and similar appearance.
Instead of utilizing pixel entities directly, superpixels can provide semantic representation and reduce the dimension of primitives \cite{ref1,ref17}, which have been widely applied in practical and industrial tasks of computer vision, such as saliency detection \cite{ref3,ref4,TII2,crm2019tcsvt,crm2020tc,DPANet,DAFNet,crm2019tip,crm2019tgrs,crm2019tmm,crm2016spl,lcy2020tc,GCPANet,icme18,nips20,eccv20,ijcai20,gcl2019tip}, feature selection \cite{TII1}, object recognition\cite{TII3}, background subtraction\cite{TIE2}, image enhancement and reconstruction\cite{TIE1,crm2016tip,gcl2020tip}.
In recent years, numerous methods with diverse problem formulations have been proposed for superpixel segmentation \cite{ref18,ref20,ref22,zhao2018flic,ASS} and achieved excellent performance in theoretical study.
However, one obvious weakness of the existing superpixel methods is the limitation to adapt to the local image details, such as the content boundaries and object contours. And in many cases of practical problems and industrial tasks \cite{ref3,ref4,TII2,TII3}, the local details are the key information, which should not be ignored. For most methods, the detailed boundaries of image content are still hard to be well preserved, due to the lack of the prior knowledge about the shape and size of the superpixels in an image. These methods have to increase the number of superpixels and reduce the superpixel size to achieve the detailed boundaries. This may usually lead to large data redundancy in the sparse area, which is unwillingness for practical tasks. How to balance the detailed information and superpixel number is a challenging problem for superpixel segmentation to be applied in industrial tasks.

To alleviate this problem, we suppose each superpixel with independent information comes from one distinct subspace, and formulate the superpixel segmentation as a subspace clustering problem \cite{8815937,SCzhang,ref32}. Subspace clustering is an effective way to segment data drawn from different dimensions or subspaces, and has achieved impressive performance in some real-world tasks, such as face clustering \cite{ref32} and foreground extraction \cite{add9}. Moreover, it also has been adopted in industrial applications, \eg, hyperspectral image classification based on superpixel-level subspace learning \cite{mei2019psasl,yu2017multiscale,li2015efficient}.
 For superpixel segmentation, each superpixel is regarded as a subspace containing pixels with similar intrinsic attributes, such as similar texture, boundary, color and coherent spatial information, while pixels in different subspace have different properties. Then the solution to the subspace clustering corresponds to the segmentation of superpixels.
 By this way, the pixels in the unit containing independent semantic information are more probably segmented into one superpixel, no matter how small the unit area is. At the same time, the size of the superpixels in sparse area are relatively large according to the similar attributes of pixels. Consequently, the detailed content boundaries and the superpixel number can be better balanced, \ie, using less superpixels to preserve more detailed information. 

Nevertheless, it is still a challenging work by inducing subspace clustering theory to superpixel segmentation directly. Different from the general sample data in clustering, pixels in an image have certain spatial correlation that only the similar pixels in adjacency tend to be clustered as one superpixel. Directly applying the conventional subspace clustering models to superpixel segmentation task can not excellently work since the spatial correlation of pixels is not taken into consideration. This may cause the confusion and segmentation error of content boundaries. In other words, we should consider the spatial constraints when using the subspace clustering for superpixel segmentation task. To this end, we devise a spatial regularization term to emphasize the spatial correlation, and propose a spatially constrained subspace clustering based superpixel segmentation model to generate superpixels with more accurate and detailed boundaries, which is more appropriate for practical and industrial tasks. 

\begin{figure*}[!t]
\centering
\includegraphics[width=0.85\textwidth]{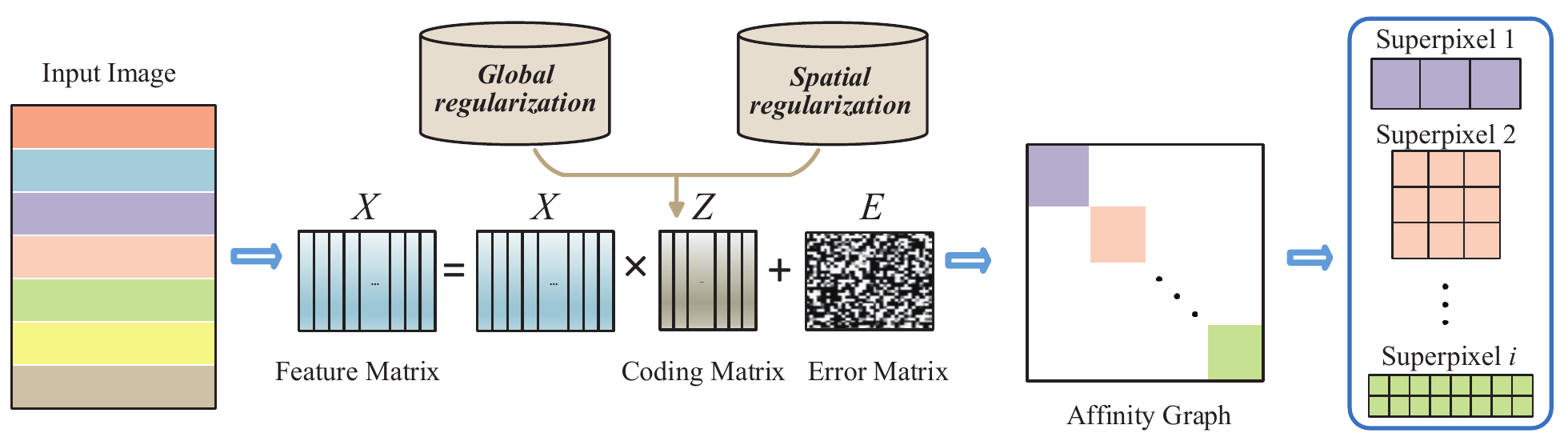}
\caption{The framework of our proposed method. Given an input image, we first construct the input data as a locality-constrained subspace clustering model with spatial regularization, and then work out the coding matrix by the proposed approach. Finally, the superpixels are generated via clustering the affinity graph of the coding matrix.}
\label{framework}
\end{figure*}


In this paper, considering the limitation of superpixel segmentation in the application of practical industrial tasks, we propose a novel spatially constrained subspace clustering based superpixel segmentation model to produce more adaptive superpixels. 
In summary, the contributions of this paper include:

\begin{itemize}
\item 
Considering the challenge of pursuing detailed boundaries, we formulate the superpixel segmentation task as a subspace clustering problem to generate content-aware supeprixels with more detailed boundaries and fewer primitives. To the best of our knowledge, this is the first attempt to induce subspace clustering theory to solve the superpixel segmentation problem, and effectively alleviate the challenge in superpixel segmentation.

\item To enforce the spatial adjacent pixels with similar attributes to be clustered into a superpixel corresponding to a subspace, we design a spatial regularization term in the loss function to constrain the local adjacency when clustering pixels. The ablation study has verified the effectiveness of the spatial regularization term.

\item The extensive experiments and discussions on the widely used benchmarks and datasets demonstrate that the proposed method achieves superior performance both quantitatively and qualitatively.
\end{itemize}

The paper is organized as follows. The related works are presented in Section \ref{section2}. In Section \ref{section3}, we propose a novel superpixel segmentation model. The comparative experimental results and analyses are presented in Section \ref{section4}. Finally, Section \ref{section6} concludes this paper.

\section{Related Work}
\label{section2}
Various superpixel segmentation algorithms have been proposed for different motivations and tasks. They can be classified into two categories: unsupervised approach and supervised approach.

\subsection{Unsupervised Superpixel Segmentation Methods}
Unsupervised approach mainly includes two categories: graph-based methods and clustering-based methods.

Normalized cut (Ncut) algorithm \cite{ref15} is one of the representative graph-based methods, which first groups pixels into several large areas by normalized cut, and then adopts K-means to segment the large areas into small units. However, it suffers from high computational complexity cost and low boundary adherence, which limit its utilization.
Considering the image topological structure, topology preserved regular superpixel (TPS) \cite{ref17} algorithm was put out to produce relatively regular superpixel. However, it suffers from low adherence to boundaries due to the demand of regular superpixel shape. In contrast to TPS, lazy random walk (LRW) \cite{ref18} algorithm achieves high object boundary adherence by iteratively optimizing an energy function based on texture measurement. Recently, approximately structural superpixels (ASS)\cite{ASS} was proposed to generate approximately structural superpixels by an asymmetrically square-wise superpixel segmentation way, which can largely reduce data amount as well as preserve image content boundaries.

Turbopixel\cite{ref19} is one of the early clustering-based methods, which adopts local image gradient and geometric flow to generate regular superpixel, but it suffers from the time consuming problem in practice. Simple linear iterative clustering (SLIC) \cite{ref20} is widely used due to its high efficiency. SLIC adopts K-means clustering approach and changes the search area into a local area rather than the global searching in K-means, which largely reduces the computation complexity. 
 As an extension of SLIC, Fast linear iterative clustering (FLIC) \cite{zhao2018flic} is an active search method which emphasizes the neighboring continuity. Linear spectral clustering (LSC) \cite{ref22} generates superpixels by mapping pixels into a high-dimensional space through a kernel function. Recently, Bayesian Adaptive Superpixel Segmentation (BASS) \cite{BASS} is proposed to generate adaptive superpixels by a Bayesian mixture model. However, the number of superpixel cannot be predefined, which is not convenient for certain tasks that demand to initialize the number manually.

\subsection{Supervised Superpixel Segmentation Methods}
Supervised superpixel segmentation algorithms are mainly based on deep-learning based approaches. Recently, deep learning based methods for superpixel segmentation have been proposed to incorporate deep learning technique with traditional segmentation. There are three representative deep learning based superpixel segmentation methods, Segmentation-aware Loss(SEAL) \cite{Tu-CVPR-2018}, Superpixel Sampling Networks(SSN) \cite{jampani18ssn}, and Superpixel Segmentation with Fully Convolutional Networks (FCN) \cite{9156320}.

SEAL adopts deep neural network and designs a segmentation-aware affinity learning method to keep weak object boundary in superpixel segmentation. SSN presents a differentiable superpixel sampling model which can be integrated into end-to-end trainable networks.
FCN algorithm adopts a simple fully connected network to predict superpixel on regular image grids. Then, on the basis of the predicted superpixel, it further develops a down-sampling/up-sampling scheme for deep networks to generate high-resolution output for intensive prediction tasks. By this way, the stereo matching network architecture can predict both superpixel and parallax.

\section{Proposed Method}
\label{section3}
In this section, we first explore a naive subspace clustering model in superpixel segmentation, and analyze the irrationality of the simple integration. Then, we propose a novel locality-constrained subspace clustering model with spatial regularization for image superpixel segmentation. An overview of the proposed method is illustrated in Fig. \ref{framework}. 

\subsection{Superpixels via Naive Subspace Clustering}
\label{section3A}

\begin{figure}[!t]
\centering
\includegraphics[width=\linewidth]{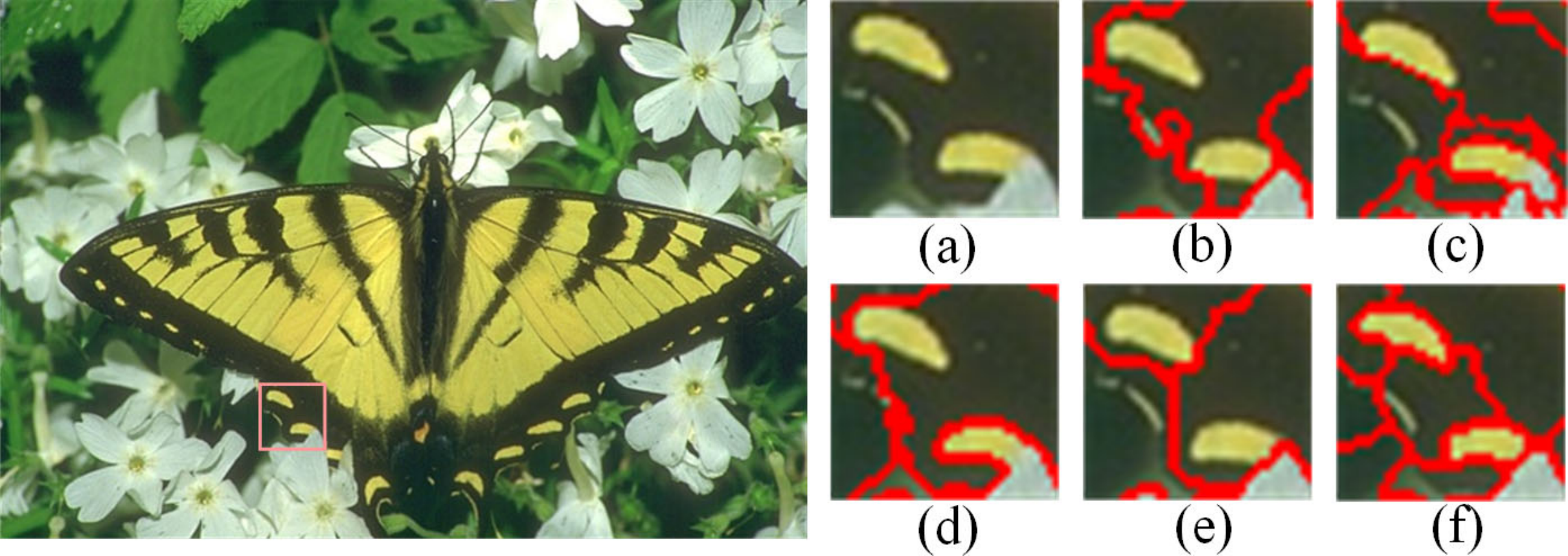}
\caption{Superpixels obtained by some state-of-the-art superpixel methods and the proposed method. (a) shows the magnified region of butterfly wing with some detailed spots. (b), (c), (d) are the results of SLIC \cite{ref20}, FLIC \cite{zhao2018flic}, and LSC \cite{ref22}, respectively. (e) shows the result based on a naive subspace clustering method. We also show the result using our spatially constrained subspace clustering method in (f).}
\label{intro}
\end{figure}
Superpixels aims at representing the original image by some units containing pixels with similar properties, such as close appearance and spatial correlation. Subspace clustering finds diverse subspaces with different dimensions and bases, and clusters each data point in one subspace \cite{ref32}. Each superpixel can be regarded as a subspace containing pixels with similar intrinsic attributes, such as similar texture, boundary, color and coherent spatial information, while pixels in different subspace have different properties. Then, superpixel segmentation problem can adopt subspace clustering theory to find the segmentation of these subspaces.

A naive way to solve this problem is to directly apply off-the-shelf subspace clustering algorithms (\eg, \cite{ref32}) to superpixel segmentation by constructing an affinity matrix by learning effective representation coding. For efficiency, we use the extracted features from the raw pixel units rather than the raw pixel features as input.
Here, a self-expressive model \cite{ref32} is adopted to learn the affinity matrix for efficiency, and the data is explained by itself: a data point can be represented as a linear combination of other data points in the same subspace, \ie, $X=XZ+E$, where $X= \left[ {{x_1},{x_2}, \cdots ,{x_n}} \right]$ is the feature matrix, $Z$ is the coding matrix, and $E$ is the error matrix. Consequently, we first formulate the superpixel segmentation problem as:
\begin{equation}\label{eq2}
\begin{array}{l}
\mathop {\min }\limits_{Z,E} \;\frac{1}{2}\left\| E \right\|_F^2 + {\lambda}{\left\| Z \right\|_0},\\
{s.t.}\;X = XZ + E,\;diag(Z) = \mathbf{0}_n,
\end{array}
\end{equation}
\noindent where ${\lambda}$ is a trade-off parameter, and $\mathbf{0}_n\in\mathbf{R}^{n\times 1}$ is an all zeros vector. The constraint $diag(Z) = \mathbf{0}_n$ is used to eliminate the trivial solution, \ie, Z equals to an identity matrix.
The term $\left\| Z \right\|_0$ is a regularization to enforce the sparsity, representing the global subspace structure in $X$. $\ell_0$-norm means to find the sparsest solution. However, minimizing (\ref{eq2}) is extremely difficult since it corresponds to an NP-hard problem. To find a nontrivial sparse representation efficiently, the $\ell_0$-norm is usually relaxed to its convex surrogate, \ie, the $\ell_1$-norm \cite{ref32}.


Nevertheless, directly applying the conventional subspace clustering models to superpixel segmentation task does not work well. Fig. \ref{intro} shows the segmentation results of some state-of-the-art superpixel methods. From Fig. \ref{intro}(b)-(d), we can see that the spots in the butterfly wing are almost unrecognizable. In Fig. \ref{intro}(e), we adopt the subspace clustering model in (\ref{eq2}) to generate superpixels, and the boundaries of the spots are also not achieved. The reason is that different from the general sample data in clustering, image pixels have inherent spatial correlations, which is not taken in the consideration of the conventional models. Therefore, we design a locality constraint to pursue the spatial correlation.


\begin{figure}[!t]
\centering
\includegraphics[width=2.9in]{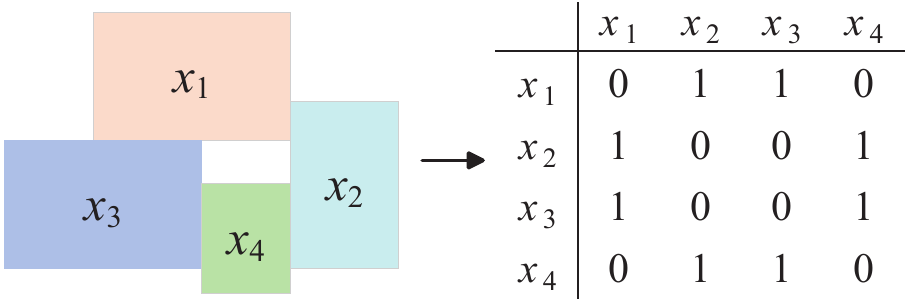}
\caption{A simple case of adjacent matrix. If two units are neighbors, the corresponding element is labeled as ``1" in the adjacent matrix; otherwise, the element is labeled as ``0".}
\label{case}
\end{figure}

\subsection{Superpixels via Locality-Constrained Subspace Clustering}
\label{section3BB}

 Considering the spatial correlation of pixels in an image that only the similar pixels in adjacency tend to be clustered as one superpixel, we devise a spatial regularization term to enforce the adjacent correlation. Then, we propose a locality-constrained subspace clustering model to generate content-aware superpixels with more detailed boundaries.


\textbf{Spatial Regularization:}
Considering a data set $X = \left[ {{x_1},{x_2}, \cdots ,{x_n}} \right]$ drawn from input image units, for image superpixel segmentation, we devise a spatial regularization term $f\left( Z \right)$ to incorporate the neighbor relationship information in $X$.  The neighbors of the $i$-th column $x_i$ (\eg, $x_{i-1},x_{i+1}$) could be similar to $x_i$. In other words, in the coding space, the corresponding neighbors of $z_i$, which is a new representation for $x_i$, could be close to $z_i$. The spatial regularization term $f\left( Z \right)$ is defined as:
\begin{equation}\label{eq5}
f\left( Z \right) = {\left\| {ZW} \right\|_{2,1}}.
\end{equation}
\noindent $\ell_{2,1}$-norm adopted here aims to achieve the similarity between neighboring pixels. $W$ is the weight matrix enforcing the neighboring correlations in $X$, which is defined as:
\begin{equation}\label{eqW}
 W = \widetilde \omega  - D,
\end{equation}
\noindent where $\omega$ is the spatial adjacent matrix of $X$, $\widetilde \omega $ is the lower triangular matrix of $\omega$, and ${D_{jj}}{\rm{ = }}\sum\limits_{i = 1}^n {{{\widetilde \omega }_{ij}}}$.

To show the spatial regularization term more clearly, denoted the adjacent relationship as $\omega$, we illustrate a simple case in Fig. \ref{case}. Specifically, $W$ is induced as follows:
$\omega {\rm{ = }}\left( {\begin{array}{*{20}{c}}
0&1&1&0\\
1&0&0&1\\
1&0&0&1\\
0&1&1&0
\end{array}} \right)$,
$\widetilde \omega {\rm{ = }}\left( {\begin{array}{*{20}{c}}
0&0&0&0\\
1&0&0&0\\
1&0&0&0\\
0&1&1&0
\end{array}} \right)$,

 \vspace{\baselineskip}

$D = \left( {\begin{array}{*{20}{c}}
2&0&0&0\\
0&1&0&0\\
0&0&1&0\\
0&0&0&0
\end{array}} \right)$,
$W = \left( {\begin{array}{*{20}{c}}
{{\rm{ - }}2}&0&0&0\\
1&{{\rm{ - }}1}&0&0\\
1&0&{{\rm{ - }}1}&0\\
0&1&1&0
\end{array}} \right)$.

 \vspace{\baselineskip}
\noindent Therefore, $ZW = \left[ {{z_2} + {z_3} - 2{z_1},{z_4} - {z_2},{z_4} - {z_3},0} \right]$, which is devised to enforce the neighbors more probably to be segmented into the same superpixel.

\textbf{Locality-Constrained Subspace Clustering Model:}
Integrated with the spatial regularization in (\ref{eq5}), we propose a locality-constrained subspace model for image superpixel segmentation:
\begin{equation}\label{eq6}
\begin{array}{l}
\mathop {\min }\limits_Z \frac{1}{2}\left\| {X - XZ} \right\|_F^2 + {\lambda _1}{\left\| Z \right\|_1} + {\lambda _2}{\left\| {ZW} \right\|_{2,1}}\;,\\
s.t.\;\;diag\left( Z \right) = {{\bf{0}}_n}\;.
\end{array}
\end{equation}
\noindent where ${\lambda _1}$ and ${\lambda _2}$ are the trade-off parameters to balance the weights of global regularization and spatial regularization, respectively.

After working out the coding matrix $Z$ (See the detailed optimization algorithm in Section \ref{section3B}), an affinity graph $G$ can be constructed for segmenting the coding matrix $Z$ and determining the subspace clusters. Similar to \cite{ref32}, the affinity graph $G$ is defined as:
\begin{equation}\label{eq21}
G{\rm{ = }}\frac{{\left| {{Z^{\rm T}}} \right|{\rm{ + }}\left| Z \right|}}{2}.
\end{equation}
\noindent This symmetrization transformation aims to assure the two connected points are in each other's sparse representation, to achieve the ideal correspondence of each element to the points from the same subspace. Then, Ncut \cite{ref15} is adopted to generate the superpixel clusters with $G$.

Moreover, in image superpixel segmentation, there may exist some isolated pixels not belonging to any clusters after the clustering step. Generally, such pixels are enforced to be assigned to the nearest cluster.
The complete procedure of the proposed method is summarized in Algorithm \ref{alg:Framwork}. We show the result using our locality-constrained subspace clustering method in Fig. \ref{intro}(f). It is clear that compared with other methods, only our method preserves the boundaries of the spots. More experimental results are presented in Section \ref{section4}.

\begin{algorithm}[!t]
\caption{Locality-Constrained Subspace Clustering Superpixel}
\label{alg:Framwork}
\textbf{Input}:  An image $I$, desired number of superpixels $K$, initial number of pixel units $n$

\textbf{Output}:  $K$ superpixels

\begin{algorithmic}[1]

\STATE Initially segment the image $I$ into $n$ raw pixel units by K-means.

\STATE Extract features of the raw pixel units as input data set $\mathbf{X} = \left[ {{\mathbf{x}_1},{\mathbf{x}_2}, \cdots ,{\mathbf{x}_n}} \right]$ (each column is the feature vector of the corresponding pixel unit).

\STATE Construct matrices $\widetilde \omega $, $D$ and $W$.

\STATE Solve the coding matrix $Z$ according to Algorithm \ref{alg:Optimization}.

\STATE Build the affinity graph $G$ with respect to $Z$ according to (\ref{eq21}).
\STATE Segment $G$ with Ncut to generate $K$ clusters.
\STATE Merge isolated pixels with the nearest neighbor clusters and update $K$.

\end{algorithmic}
\end{algorithm}

\subsection{Optimization} 
\label{section3B}
To solve the objective function in (\ref{eq6}), we adopt alternating direction method of multipliers (ADMM)\cite{ref34} to devise an optimization algorithm. Two auxiliary variables $U = Z$ and $V = UW$ are induced to separate the terms of variable $Z$. Hence, (\ref{eq6}) is equivalent to be written as:
\begin{equation}\label{eqeq6}
\begin{array}{l}
\mathop {\min }\limits_{Z,U,V} \frac{1}{2}\left\| X - XU \right\|_F^2{\rm{ + }}{\lambda _1}{\left\| Z \right\|_1}{\rm{ + }}{\lambda _2}{\left\| {V} \right\|_{2,1}}\\
{\rm{s}}{\rm{.t}}{\rm{.}}\;\;diag(Z) = \mathbf{0}_n, U=Z, V=UW.
\end{array}
\end{equation}

\noindent The augmented Lagrangian function of (\ref{eqeq6}) is:
\begin{equation}\label{eq8}
\begin{array}{l}
\mathcal{L}\left( {Z,U,V} \right) = \frac{1}{2}\left\| {X - XU} \right\|_F^2 + {\lambda _1}{\left\| Z \right\|_1} + {\lambda _2}{\left\| V \right\|_{2,1}}\\
\;\;\;\;\;\;\;\;\;\;\;\;\;\;\;\;\;\;\;\; + \left\langle {\Theta ,Z - U} \right\rangle  + \frac{{{\alpha _1}}}{2}\left\| {Z - U} \right\|_F^2\\
\;\;\;\;\;\;\;\;\;\;\;\;\;\;\;\;\;\;\;\; + \left\langle {\Xi ,V - UW} \right\rangle  + \frac{{{\alpha _2}}}{2}\left\| {V - UW} \right\|_F^2
\end{array}
\end{equation}
\noindent where $\Theta\in\mathbf{R}^{n\times n}$ and $\Xi\in\mathbf{R}^{n\times n}$ are the Lagrangian multiplier matrices, ${\alpha _1}$ and ${\alpha _2}$ are the regularization parameters.

We can solve for $U$, $V$ and $Z$ alternatively by minimizing $\mathcal{L}$ when fixing the other two.

\subsubsection{\textbf{Update Z when fixing U and V}}
The problem in (\ref{eq8}) becomes
%
\begin{equation}\label{eq10}
\mathop {\min }\limits_Z {\lambda _1}||Z|{|_1} + \frac{{{\alpha _1}}}{2}\left\|Z - \left(U - \frac{\Theta }{{{\alpha _1}}}\right)\right\|_F^2.
\end{equation}
\noindent Since (\ref{eq10}) is element-wise to each $Z_{ij}$, by soft thresholding operator \cite{ref40}, we can get a closed-form solution:
\begin{equation}\label{apeq11}
Z = {\rm{sign}}\left( U - \frac{\Theta }{{{\alpha _1}}}\right) {\rm{max}}\left( \left| {U - \frac{\Theta }{{{\alpha _1}}}} \right| - \frac{{{\lambda _1}}}{{{\alpha _1}}}\right).
\end{equation}




\subsubsection{\textbf{Update U when fixing Z and V}}
\begin{equation}\label{eq12}
\begin{array}{l}
\mathop {\min }\limits_U \frac{1}{2}\left\| {X - XU} \right\|_F^2 + \left\langle {\Theta ,Z - U} \right\rangle  + \frac{{{\alpha _1}}}{2}\left\| {Z - U} \right\|_F^2\\
\;\;\;\;\;\;\; + \left\langle {\Xi ,V - UW} \right\rangle  + \frac{{{\alpha _2}}}{2}\left\| {V - UW} \right\|_F^2
\end{array}
\end{equation}
\noindent By setting the derivative with respect to $U$ to zero, we can obtain the following equation:
\begin{equation}\label{eq13}
\begin{array}{l}
\left( {{X^{\rm T}}X + {\alpha _1}I} \right)U + {\alpha _2}UW{W^{\rm T}}\\
 = {X^{\rm T}}X + {\alpha _2}V{W^{\rm T}} + {\alpha _1}Z + \Theta  + \Xi {W^{\rm T}}.
\end{array}
\end{equation}
\noindent (\ref{eq13}) is a standard Sylvester equation \cite{ref41}, which can be linearized and solved by:
\begin{equation}\label{apeq14}
\begin{array}{l}
[I \otimes ({X^{\rm T}}X + {\alpha _1}I) + {\alpha _2}W{W^{\rm T}} \otimes I]{\rm{vec(}}U{\rm{)}}\\
{\rm{ = vec}}({X^{\rm T}}X + {\alpha _2}V{W^{\rm T}} + {\alpha _1}Z + \Theta  + \Xi {W^{\rm T}}),
\end{array}
\end{equation}
\noindent where $ \otimes $ is the tensor product and vec$(\cdot)$ returns a column vector by stacking the columns of a matrix.




\subsubsection{\textbf{Update V when fixing Z and U}}
\begin{equation}\label{eq15}
\mathop {\min }\limits_V {\lambda _2}{\left\| V \right\|_{2,1}} + \left\langle {\Xi ,V - UW} \right\rangle  + \frac{{{\alpha _2}}}{2}\left\| {V - UW} \right\|_F^2.
\end{equation}
\noindent (\ref{eq15}) is equivalent to
\begin{equation}\label{eq16}
\mathop {\min }\limits_V {\lambda _2}{\left\| V \right\|_{2,1}} + \frac{{{\alpha _2}}}{2}\left\|V - \left(UW - \frac{1}{{{\alpha _2}}}\Xi \right)\right\|_F^2.
\end{equation}
\noindent Let $N = UW - \frac{1}{{{\alpha _2}}}\Xi $, (\ref{eq16}) can be solved by a closed-form solution \cite{ref42}:
\begin{equation}\label{apeq17}
V(:,i) = \left\{ {\begin{array}{*{20}{c}}
{\frac{{||N(:,i)|| - \frac{{{\lambda _2}}}{{{\alpha _2}}}}}{{||N(:,i)||}}N(:,i),}&{{\rm if}\;||N(:,i)||\; > \frac{{{\lambda _2}}}{{{\alpha _2}}},}\\
{0,}&{{\rm otherwise},}
\end{array}} \right.
\end{equation}
\noindent where $V(:,i)$ denotes the $i$th column of $V$.




\subsubsection{\textbf{Update $\Theta$, $\Xi$, ${\alpha _1}$ and ${\alpha _2}$ by}}
\begin{equation}\label{eq18}
\Theta {\rm{ = }}{\Theta ^{old}} + {\alpha _1}(Z - U),
\end{equation}
\begin{equation}\label{eq19}
\Xi {\rm{ = }}{\Xi ^{old}} + {\alpha _2}(V - UW),
\end{equation}
\begin{equation}\label{eq20}
\alpha_1=min(\rho\alpha_1^{old},\alpha_1^{max});\alpha_2=min(\rho\alpha_2^{old},\alpha_2^{max}),
\end{equation}
\noindent where $\Theta\in\mathbf{R}^{n\times n}$ and $\Xi\in\mathbf{R}^{n\times n}$ are initialized as all zeros matrix, $\rho=1.1$, $\alpha _1$ and $\alpha _2$ are both initialized as 0.01, $\alpha_1^{max}= 10^8$, $\alpha_2^{max}= 10^8$.

The complete optimization algorithm is summarized in Algorithm \ref{alg:Optimization}.

\begin{figure}[!t]
\centering
\subfloat[BSDS500]{\includegraphics[width=0.23\textwidth]{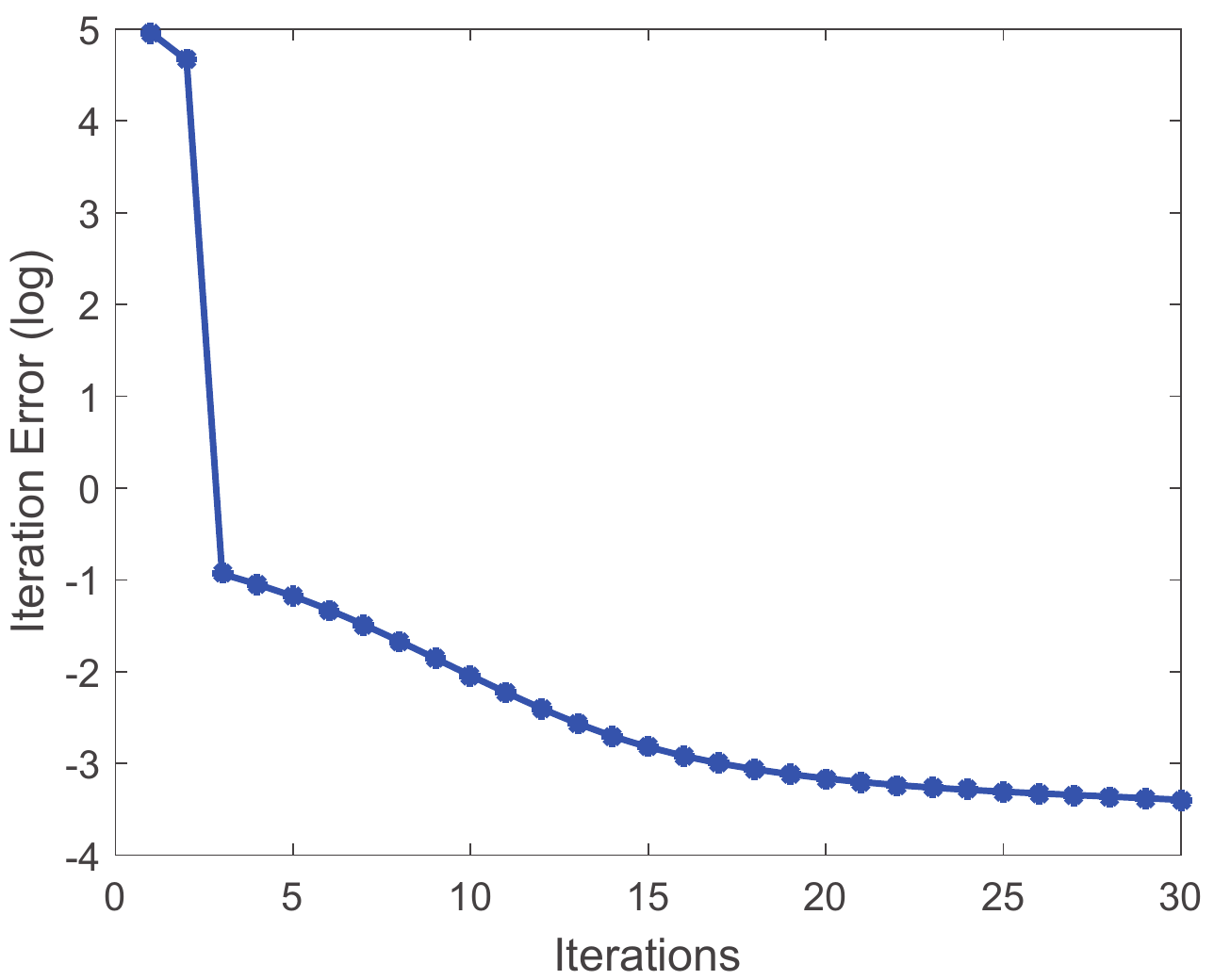}%
\label{BSDS500}}
\hfil
\subfloat[NYUv2]{\includegraphics[width=0.23\textwidth]{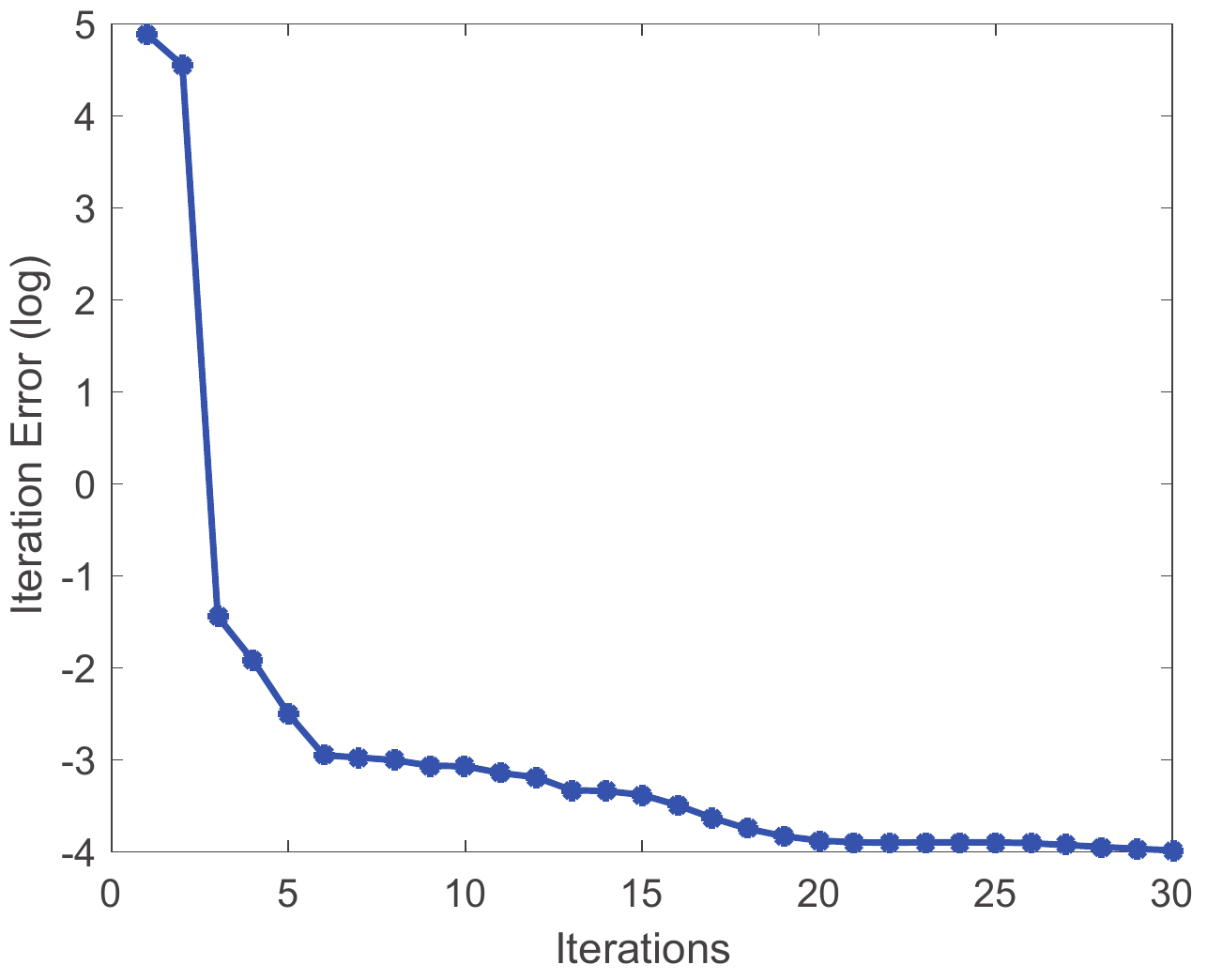}%
\label{NYUv2}}
\caption{The convergence processes of different datasets.}
\label{fig-convergence}
\end{figure}

\begin{algorithm}[!t]
\caption{Optimization to (\ref{eq6})}
\label{alg:Optimization}
\textbf{Input}:  $X$, $W$, maximum iterations $iter_{max}$, parameters ${\lambda _1}$, ${\lambda _2}$, $i=0$

\textbf{Output}:  $Z$
\begin{algorithmic}[1]

\WHILE {$ not \ converged $ and $i \le iter_{max}$}
\STATE ${Z_{i + 1}} = \arg \mathop {\min }\limits_Z \mathcal{L}({Z_i},{U_i},{V_i})$;(referred to (\ref{apeq11}))
\STATE ${U_{i + 1}} = \arg \mathop {\min }\limits_U \mathcal{L}({Z_i},{U_i},{V_i})$; (referred to (\ref{apeq14}))
\STATE ${V_{i + 1}} = \arg \mathop {\min }\limits_V \mathcal{L}({Z_i},{U_i},{V_i})$; (referred to (\ref{apeq17}))
\STATE Update $\Theta$, $\Xi$, $\alpha _1$ and $\alpha _2$ according to (\ref{eq18})-(\ref{eq20});
\STATE $i=i+1$;
\ENDWHILE

\end{algorithmic}
\end{algorithm}

\subsection{Discussion}
In the proposed model, three blocks of variables $Z$, $V$, and $U$ need to be optimized in eq. (6). In each iteration, with the help of the ADMM, the objective function can converge to a local optimum with the optimized variables. Though there is no general convergence proof for the ADMM based algorithm with more than two blocks of variables \cite{7368899}, to the best of our knowledge, we observe that our algorithm empirically converges well and produces satisfactory performance on real-world data sets, contributing to the fact that the proposed model is convex, and each subproblem has a closed form solution.

The convergence curve is illustrated in Fig. \ref{fig-convergence}.
The algorithm can be updated iteratively until it the stopping conditions are satisfied, \ie, the iteration error is reduced to a stable value ($10^{-3}$ in our experiments) after several iterations or the algorithm exceeds the maximum iteration. We empirically find that 20 iterations are sufficient for most images, and thus we set the maximum iteration number to 20 for efficiency.

The computational complexity of the proposed method is dominated by the initialization by K-means and the optimization process. The computational complexity of K-means is $O(N)$, where $N$ is the total number of pixels in the input image. That of the optimization process is $O(k^3)$, where $k$ is the initial number of pixel units, which is far less than the total pixel number N. Hence, the time complexity for the proposed method is $O(N+k^3)$.


\begin{figure}[!t]
\centering
        \includegraphics[width=1.65in]{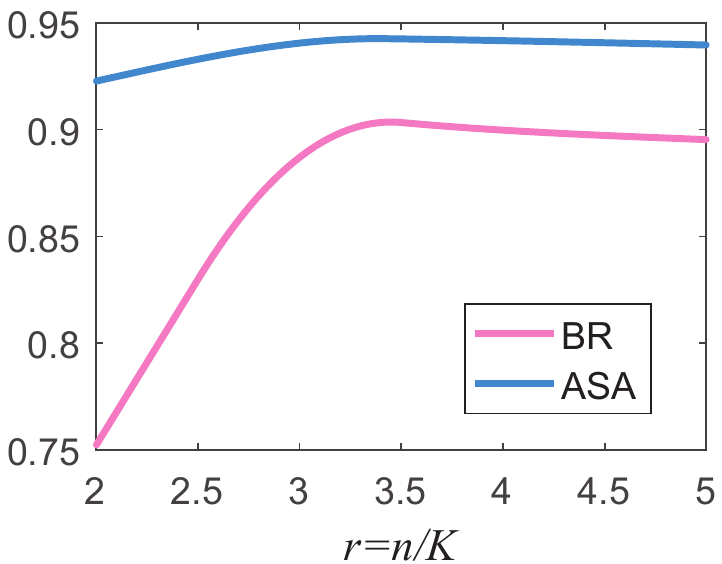}
\hfil
        \includegraphics[width=1.65in]{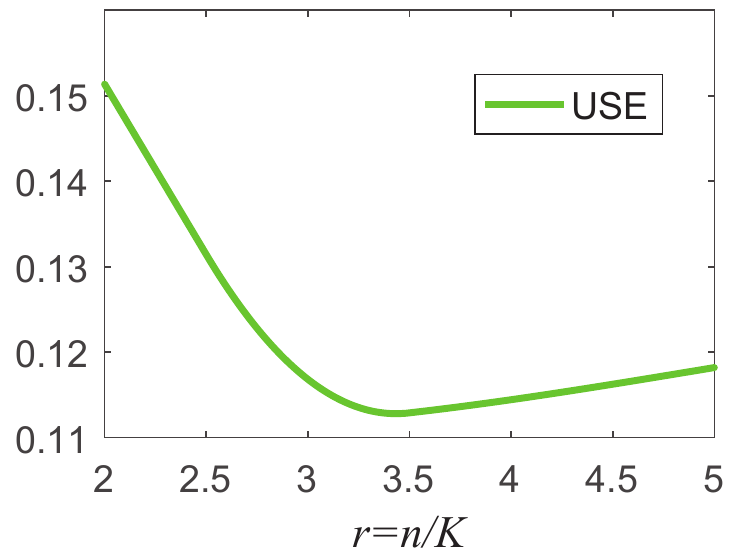}
\caption{The influence of $r$ on the segmentation results.} 
\label{fig-ratio}
\end{figure}

\section{Experimental Results and Discussions}
\label{section4}

To show the merits qualitatively and quantitatively, we compare the proposed method with some state-of-the-art methods, including Ncut \cite{ref15}, Turbopixel\cite{ref19}, TPS \cite{ref17}, LRW \cite{ref18}, SLIC\cite{ref20}, ASS \cite{ASS}, FLIC \cite{zhao2018flic}, LSC \cite{ref22} and BASS \cite{BASS}. The parameters of the compared methods are set according to the original works. Since BASS cannot set the superpixel number manually, we adopt the published results directly.


\subsection{Evaluation Criteria and Datasets}
In the experiments, three commonly used metrics are utilized to evaluate all of the methods quantitatively: achievable segmentation accuracy (ASA), boundary recall (BR) and under-segmentation error (USE)\cite{zhao2018flic}. ASA measures the highest achievable accuracy of a segmentation, \ie, the percentage of the most labeled pixels of superpixel segments overlapped from the ground truth segments with respect to the total number of the pixels in superpixel segments. Higher ASA indicates better segmentation of image content.
USE measures the fraction of pixels of superpixel segments that not overlapped from the ground truth segments. Lower USE means fewer pixels leak from the object boundaries and edges.
BR is the fraction of correctly overlapped pixels of superpixel segments from the ground truth boundaries. A high BR means the image content boundaries are well preserved after segmentation.






\begin{figure*}[!t]
\centering
\includegraphics[width=0.95\textwidth]{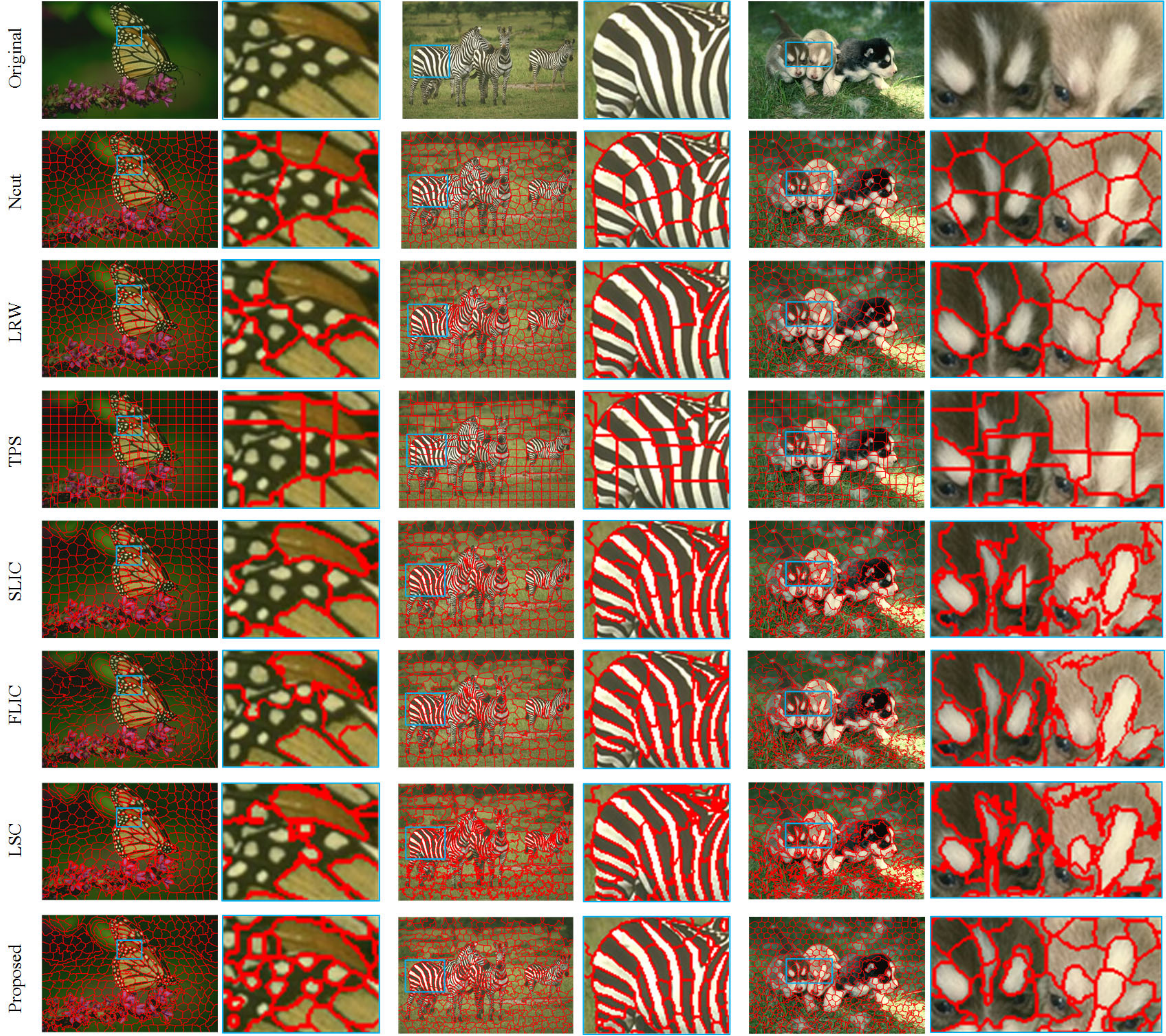}
\caption{Visual comparison with some state-of-the-art methods. The first, third, and fifth columns are the image superpixel maps. The second, fourth, and sixth columns are the corresponding magnified regions.}
\label{unsupervised}
\end{figure*}

 We compare the proposed method with some state-of-the-art superpixel segmentation methods on the public Berkeley Segmentation Dataset (BSDS500) \cite{ref44} for thorough superpixel segmentation evaluation. BSDS500 contains 500 natural images including humans, animals, buildings, landscape and outdoor scenes. Since the dataset has accurate annotated segmentation boundaries, almost all of the superpixel methods adopt BSDS500 for evaluation. The dataset provides 5 ground truth for each image and we average segmentation results with respect to the different ground truth for evaluation.
For further comparisons, we evaluate the methods on two other datasets: Densely Annotated VIdeo Segmentation (DAVIS) \cite{DAVIS2016} and Cityscapes \cite{cityscapes}. Since these datasets do not contain accurate boundaries, the ASA metric is adopted to evaluate the upper bound segmentation performance. 
Besides, we add NYUv2 \cite{NYUv2} dataset which consists of indoor-scene images for further extensive comparison.

\begin{figure*}[!t]
\centering
\subfloat[LRW: $K=1600$]{\includegraphics[width=0.21\textwidth]{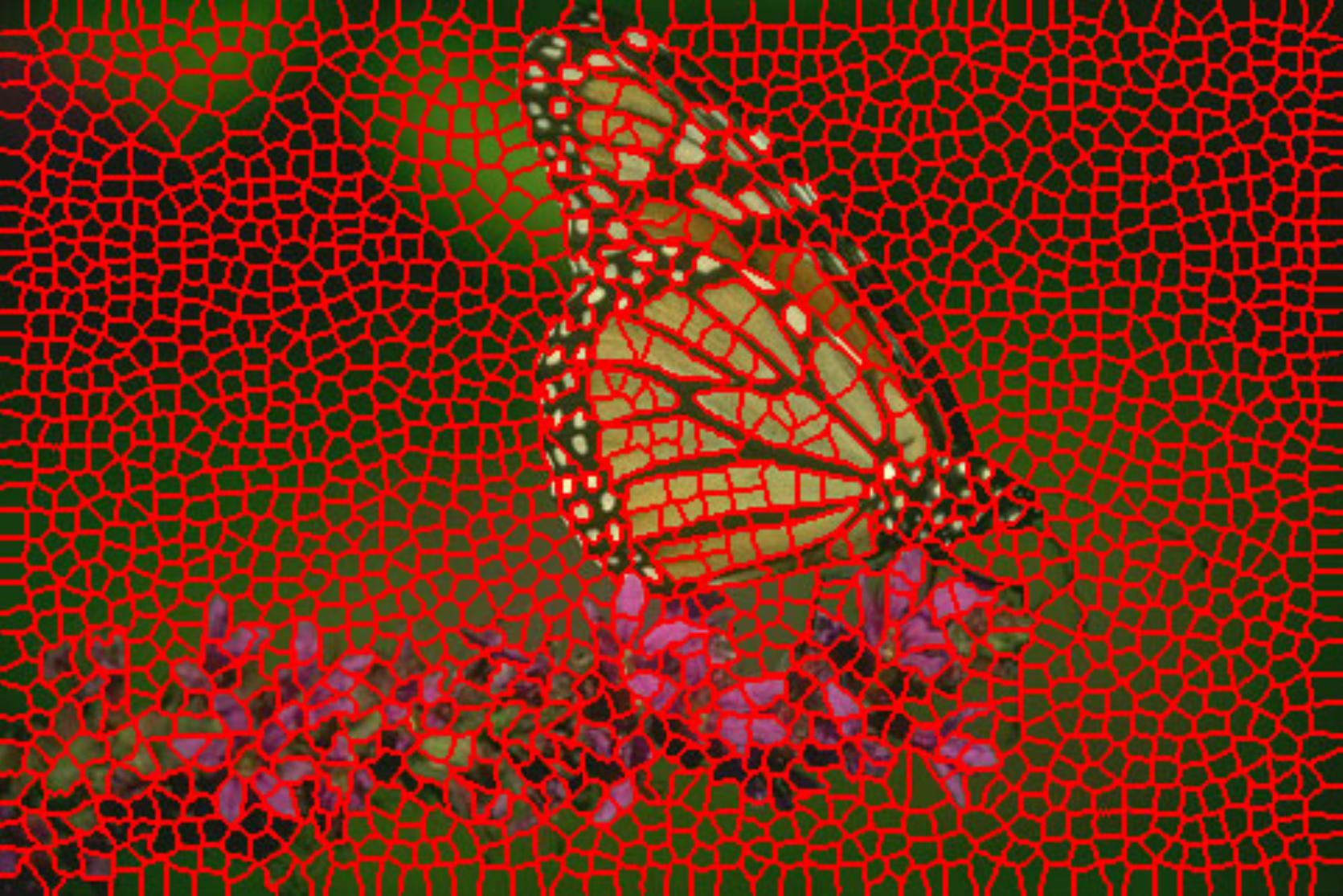}%
\label{LRW}}
\hfil
\subfloat[SLIC: $K=1800$]{\includegraphics[width=0.21\textwidth]{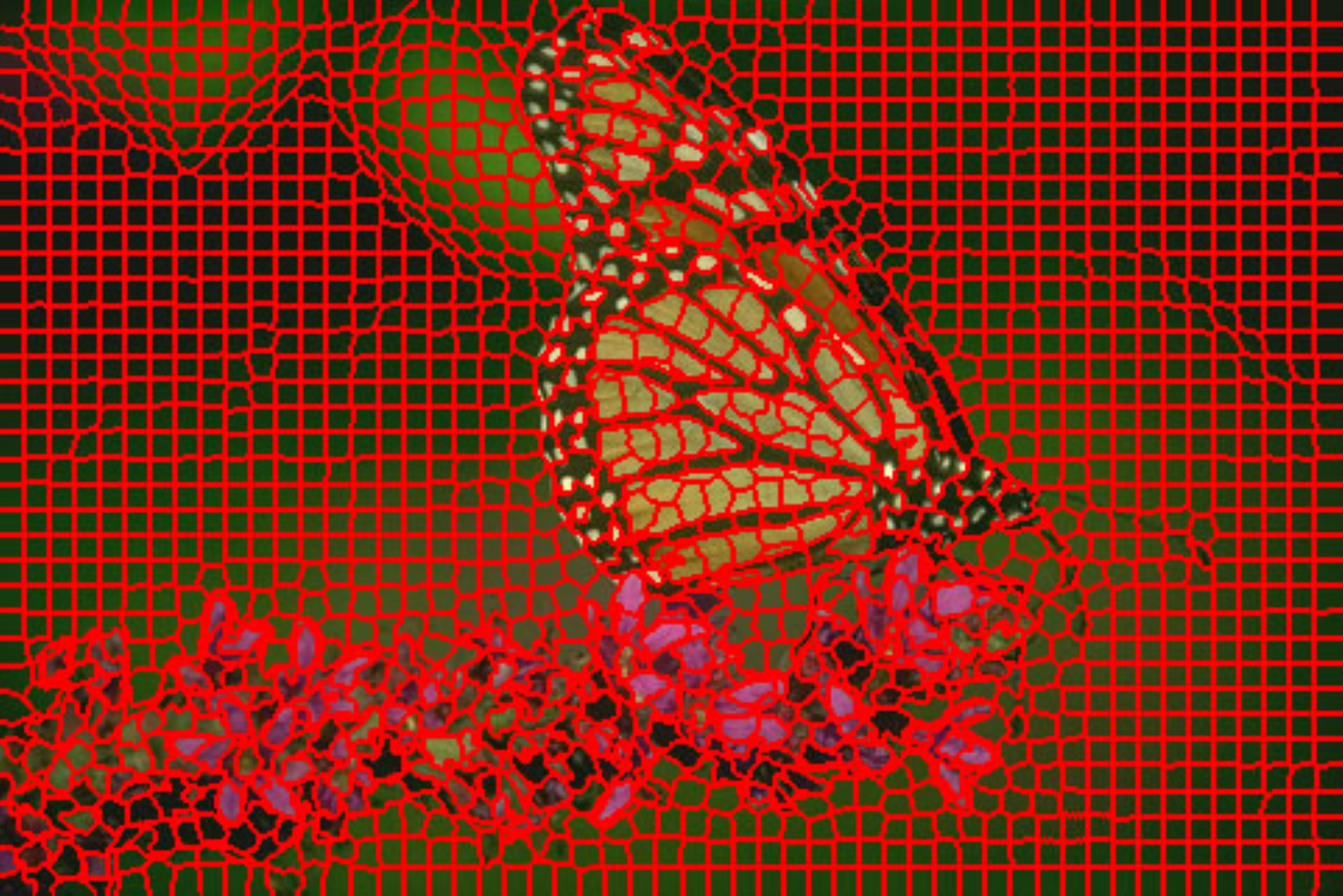}%
\label{SLIC}}
\hfil
\subfloat[LSC: $K=1600$]{\includegraphics[width=0.21\textwidth]{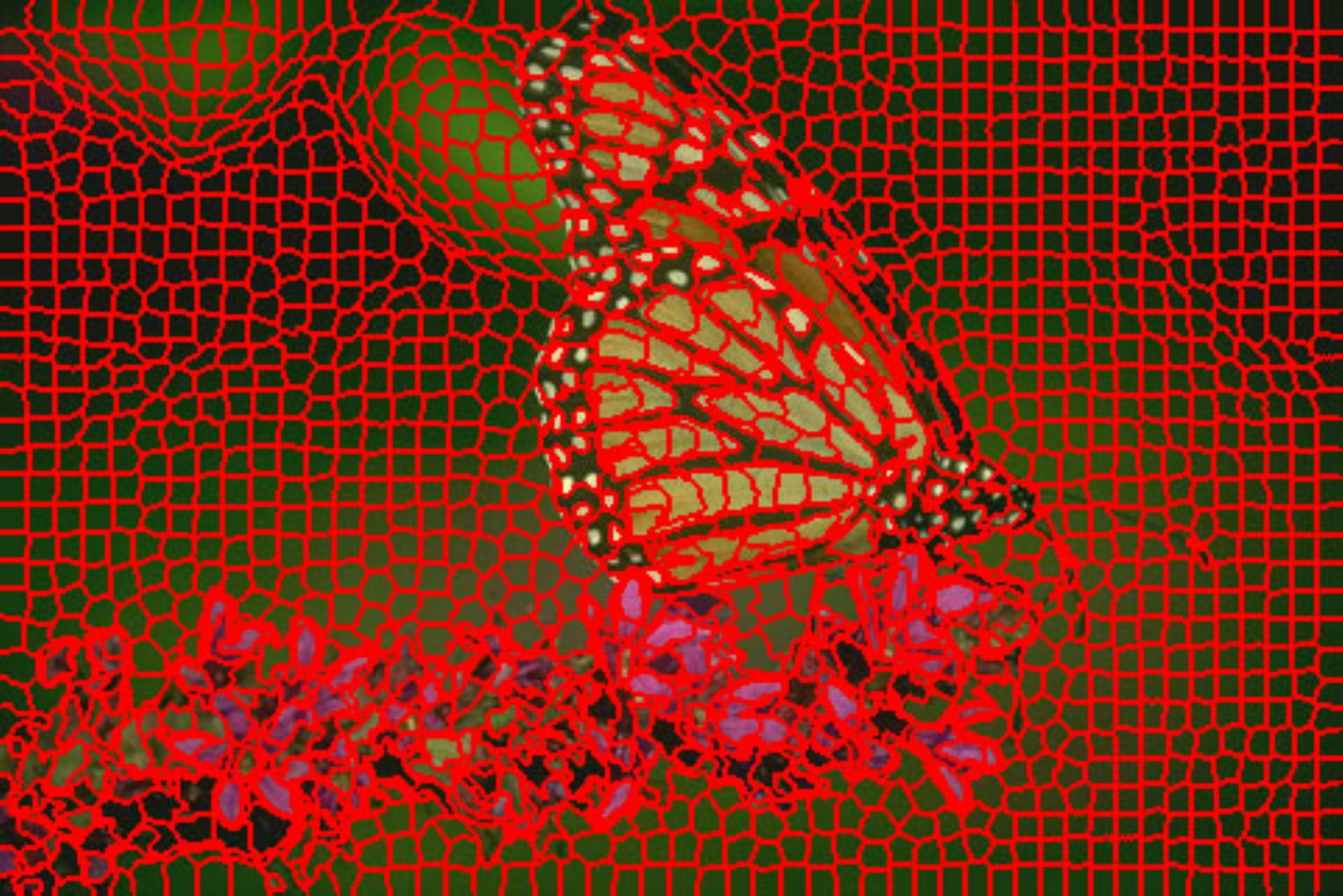}%
\label{LSC}}
\hfil
\subfloat[Proposed: $K=300$]{\includegraphics[width=0.21\textwidth]{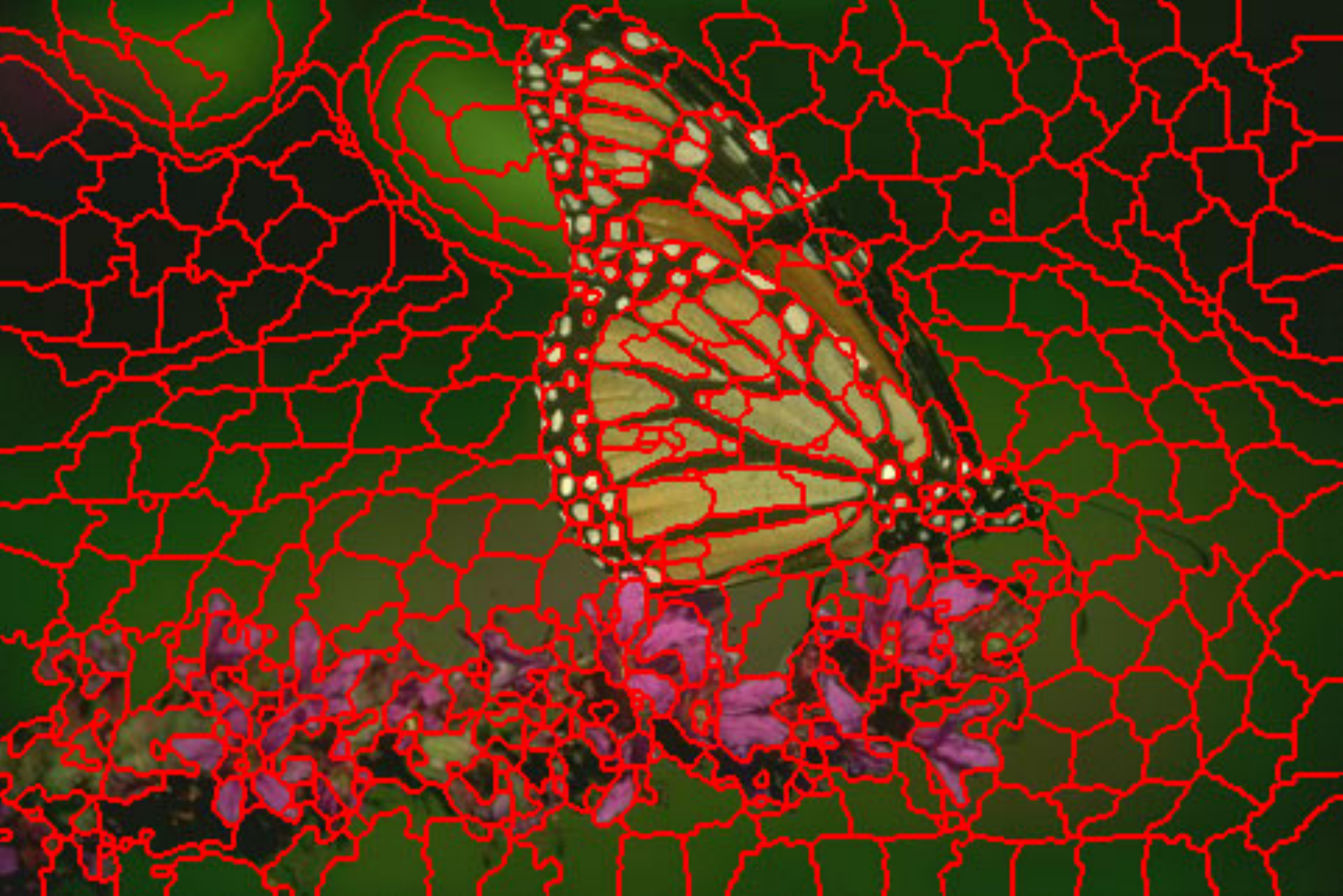}%
\label{SCS}}

\caption{We compare the minimum initial settings of the desired superpixel number $K$ for spot segmentation in the butterfly image. The proposed method can segment the spot when $K=300$, while LRW, SLIC, and LSC are 1600, 1800, and 1600, respectively.}
\label{dot}
\end{figure*}
\begin{figure*}[!t]
\centering
\includegraphics[width=0.9\textwidth]{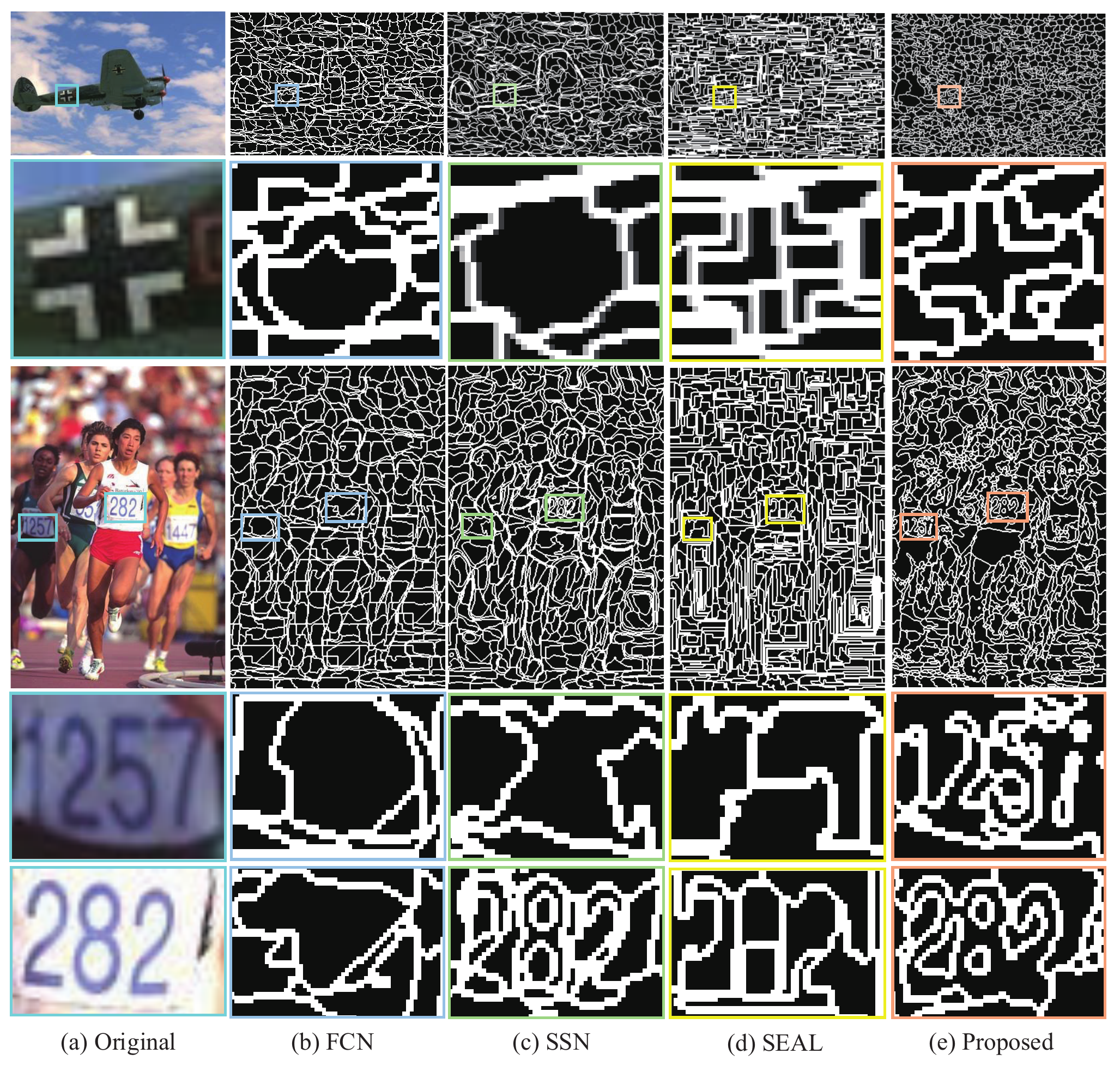}
\caption{Visual comparison with three recent deep learning based methods, FCN, SSN and SEAL. (a) is the original image and corresponding magnified regions. (b), (c) (d) and (e) are the superpixel label maps of FCN, SSN, SEAL, and the proposed method, respectively.}
\label{SCS-deep}
\end{figure*}


\begin{figure*}[!t]
\centering
\subfloat[BSDS500]{\includegraphics[width=0.3\textwidth]{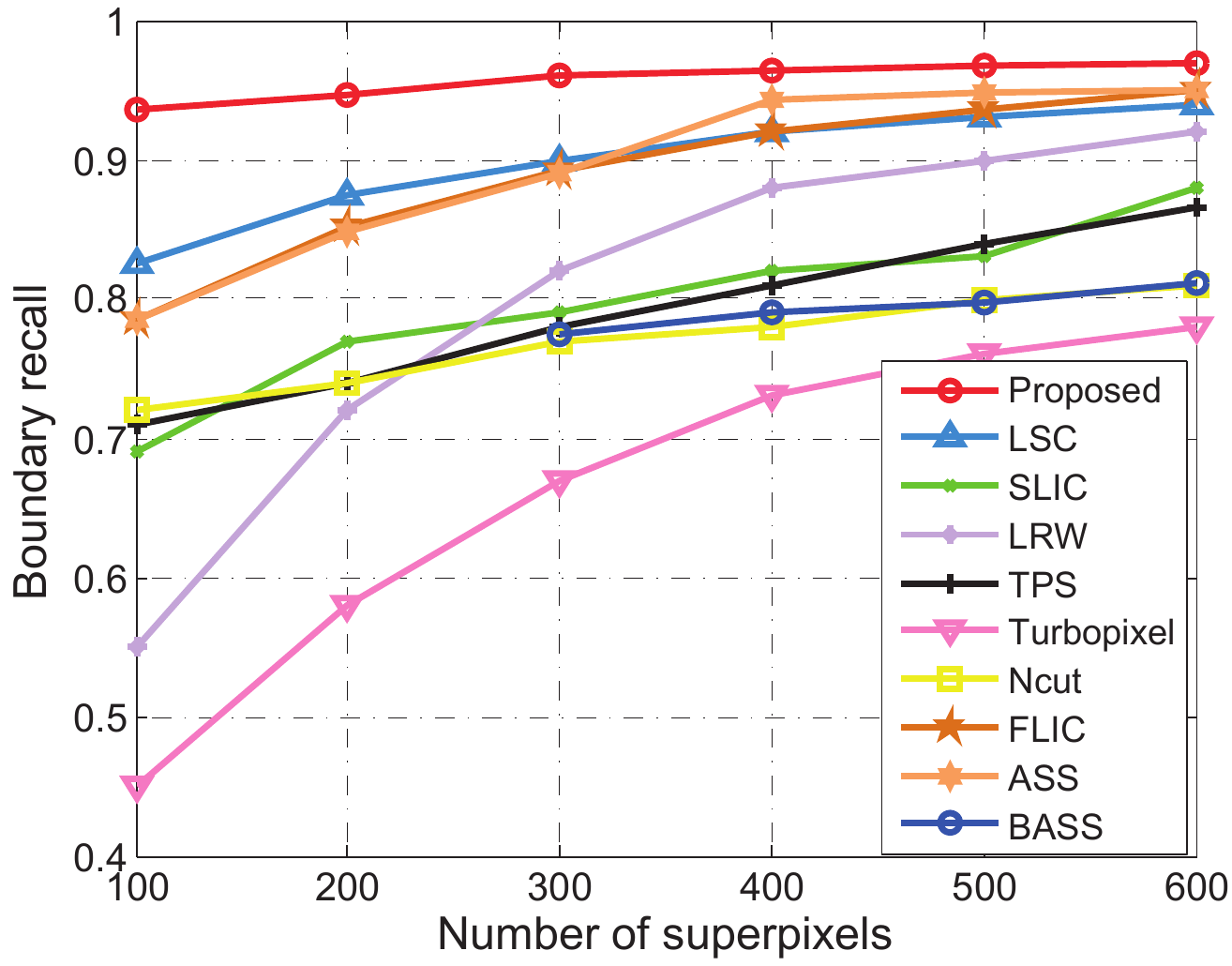}%
\label{BR}}
\hfil
\subfloat[BSDS500]{\includegraphics[width=0.3\textwidth]{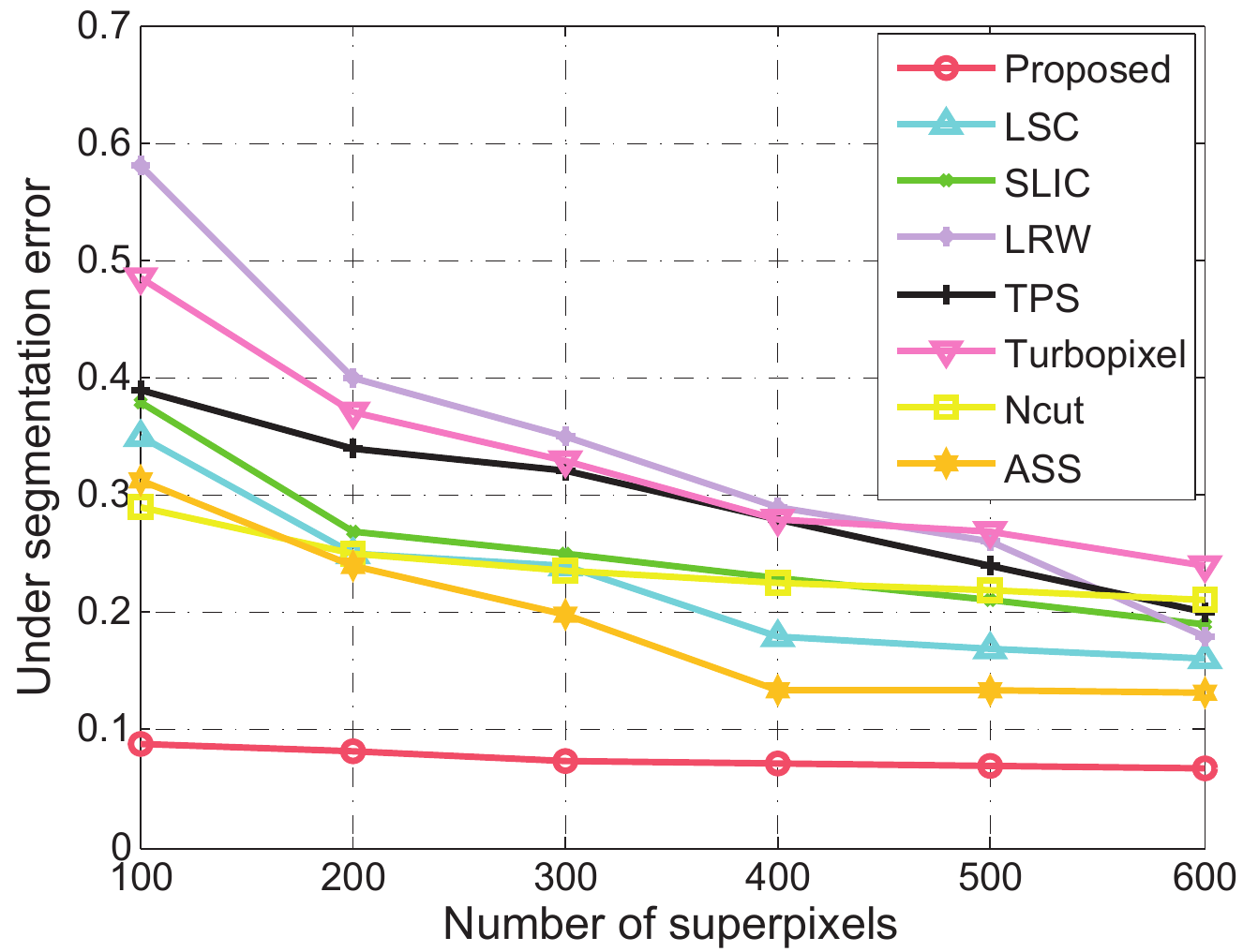}%
\label{USE}}
\hfil
\subfloat[BSDS500]{\includegraphics[width=0.3\textwidth]{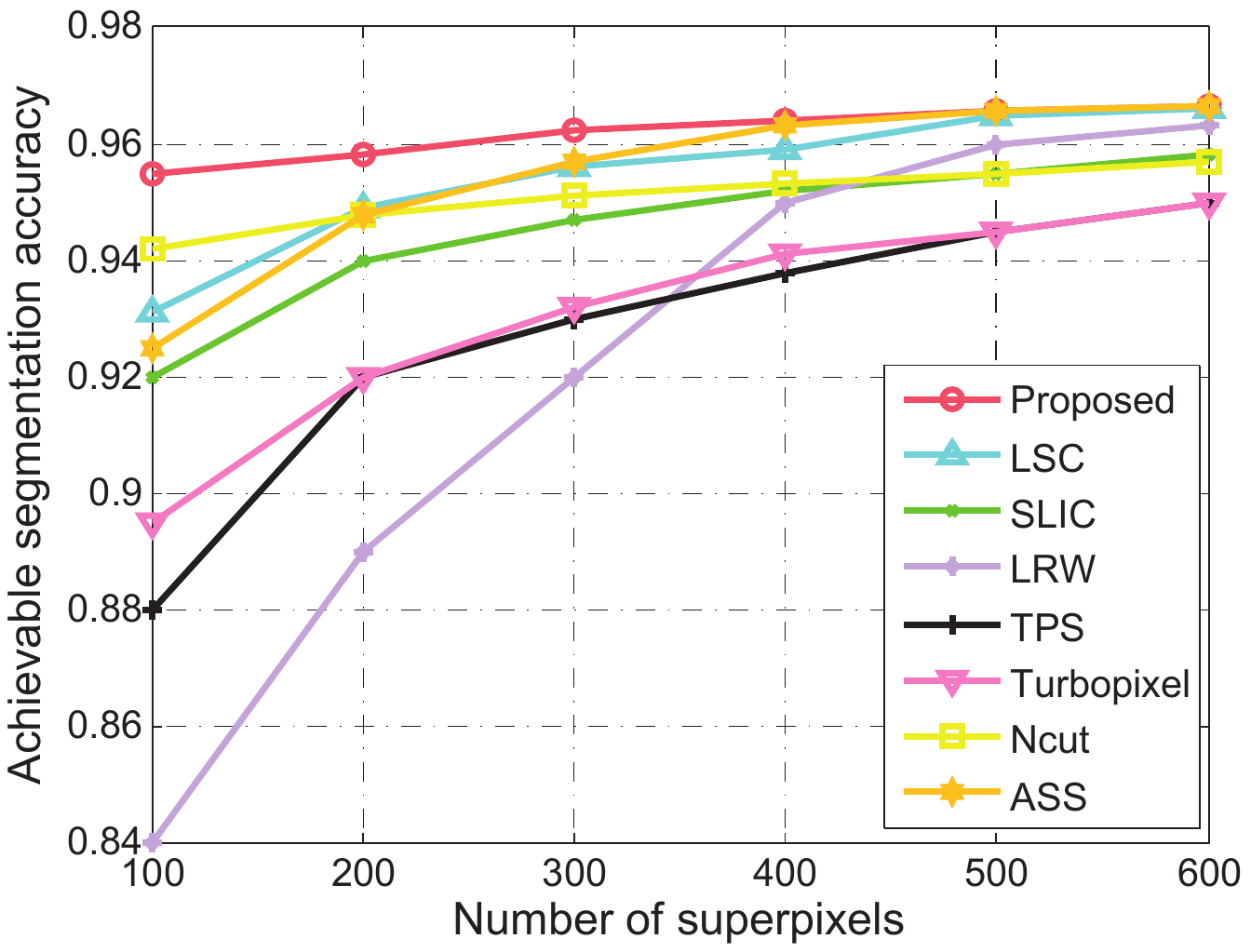}%
\label{ASA}}
\hfil
\subfloat[DAVIS]{\includegraphics[width=0.3\textwidth]{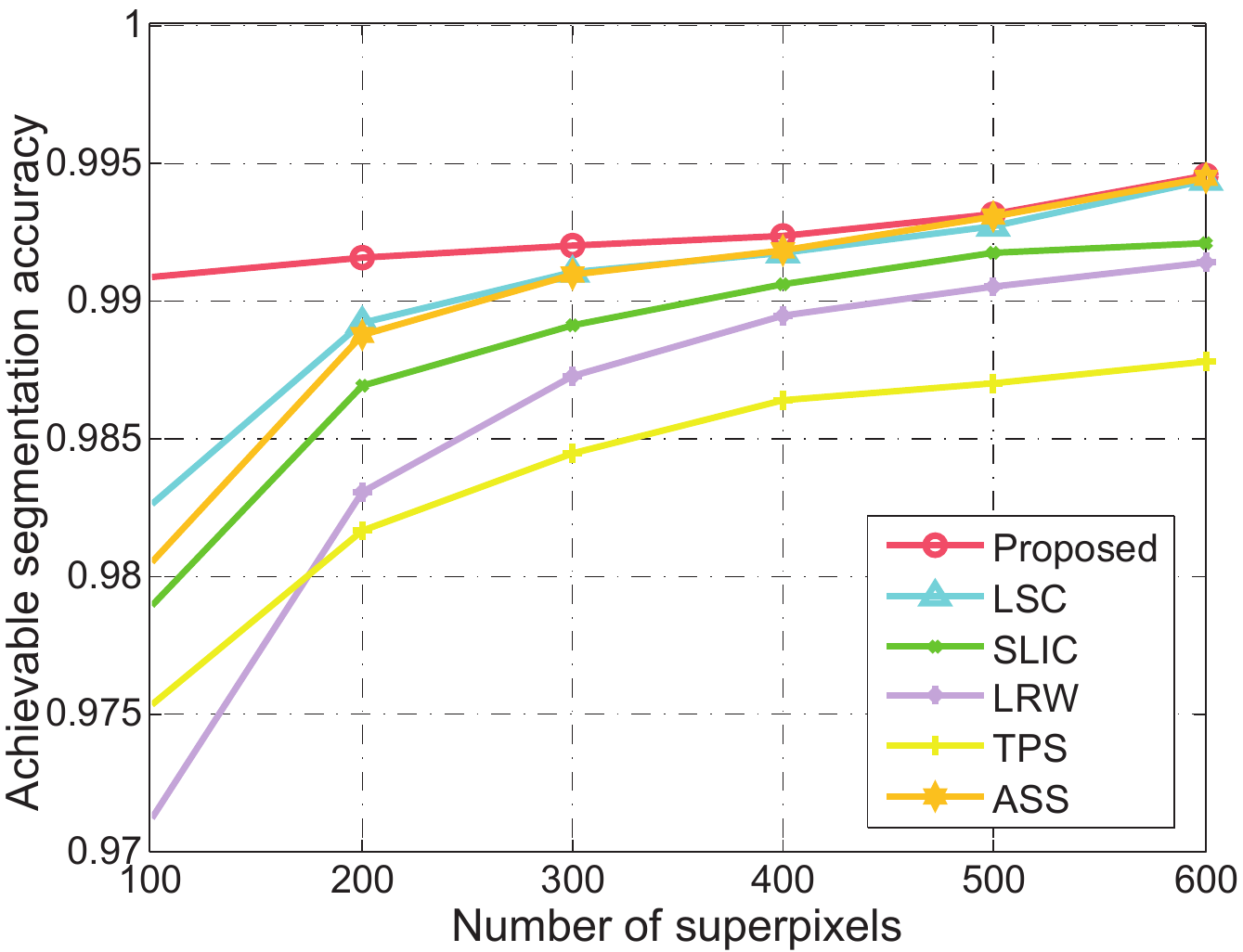}%
\label{ASA-DAVIS}}
\hfil
\subfloat[Cityscapes]{\includegraphics[width=0.3\textwidth]{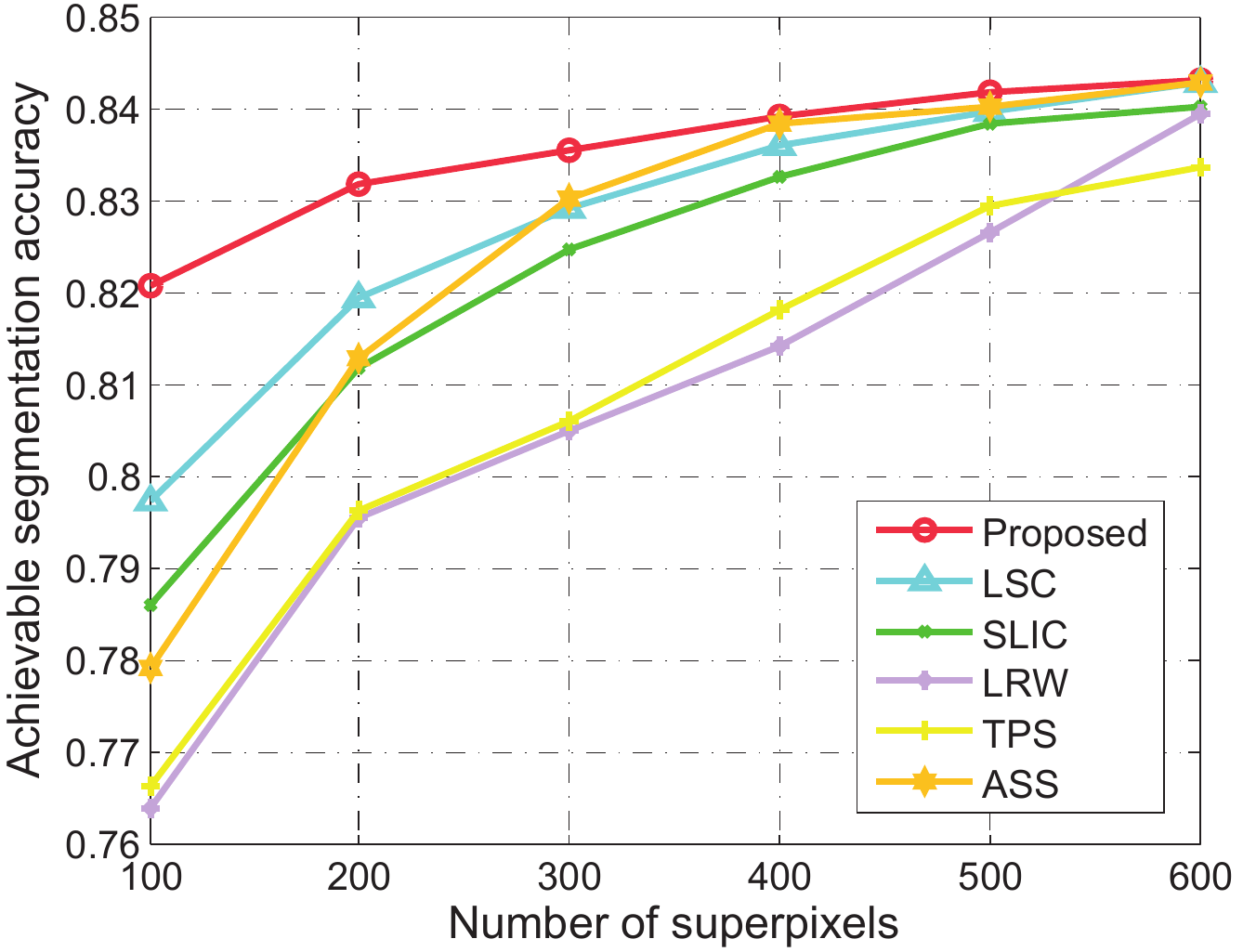}%
\label{ASA-Cityscapes}}
\hfil
\subfloat[BSDS500]{\includegraphics[width=0.3\textwidth]{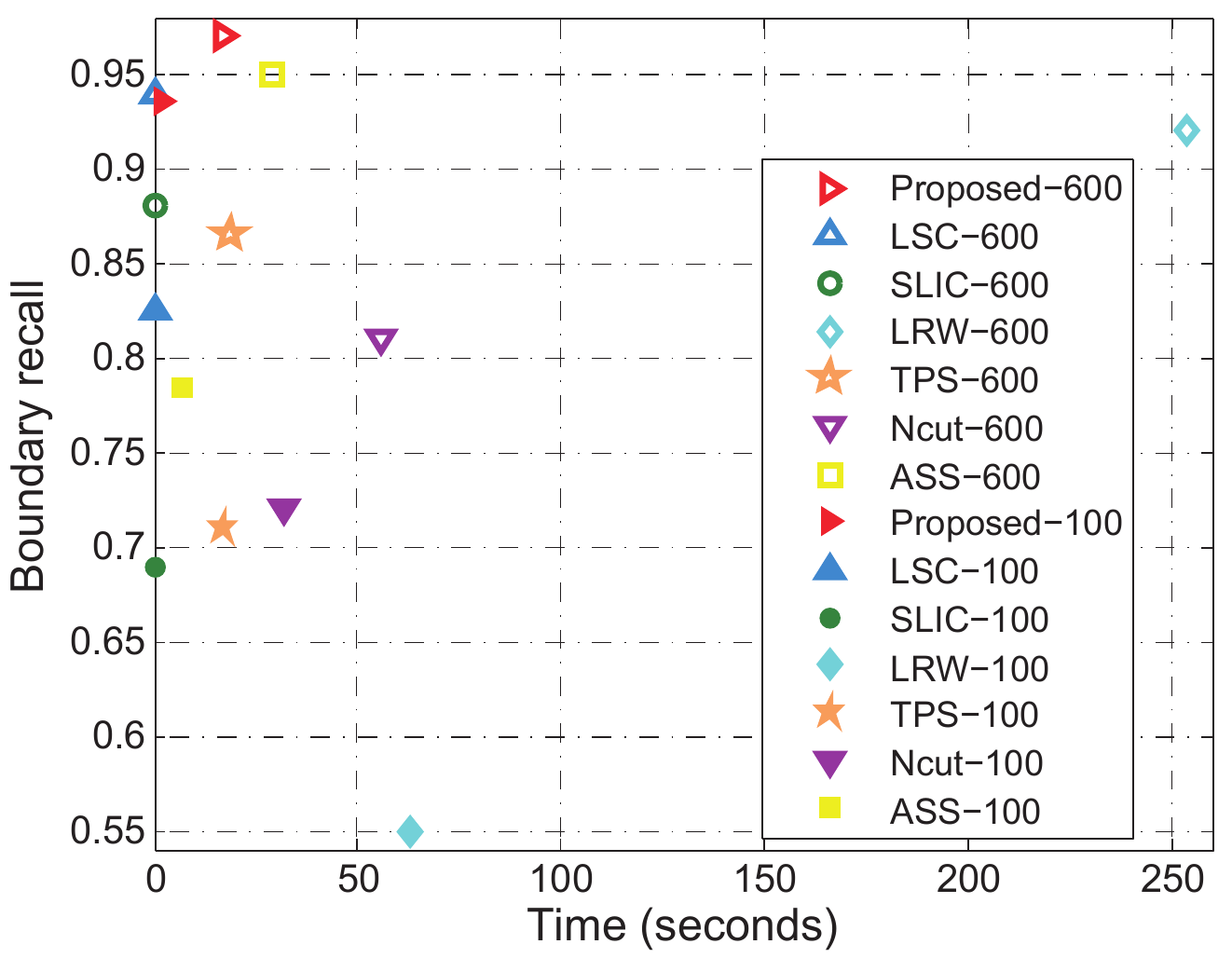}%
\label{BR-time}}
\hfil
\subfloat[NYUv2]{\includegraphics[width=0.3\textwidth]{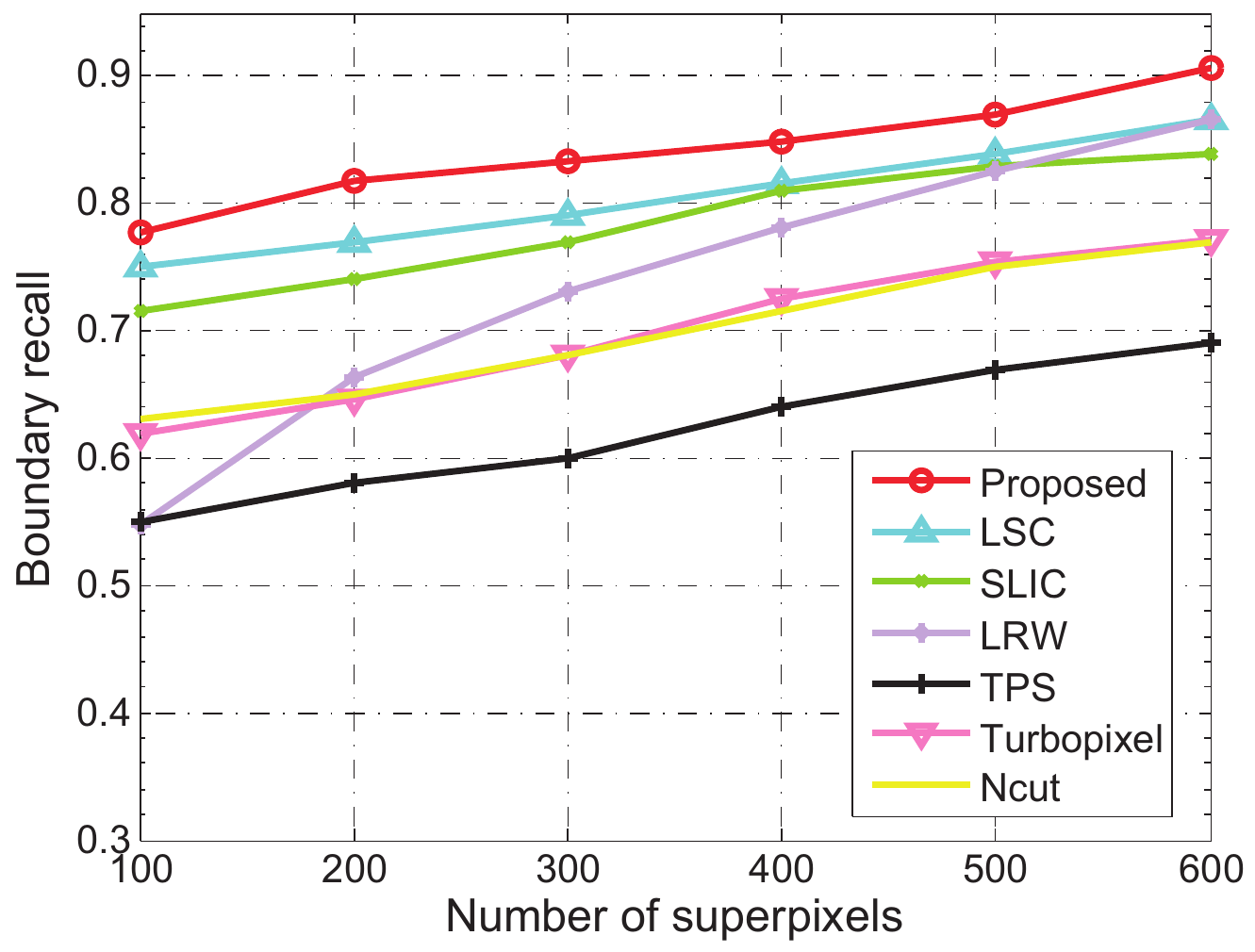}%
\label{BR-N}}
\hfil
\subfloat[NYUv2]{\includegraphics[width=0.3\textwidth]{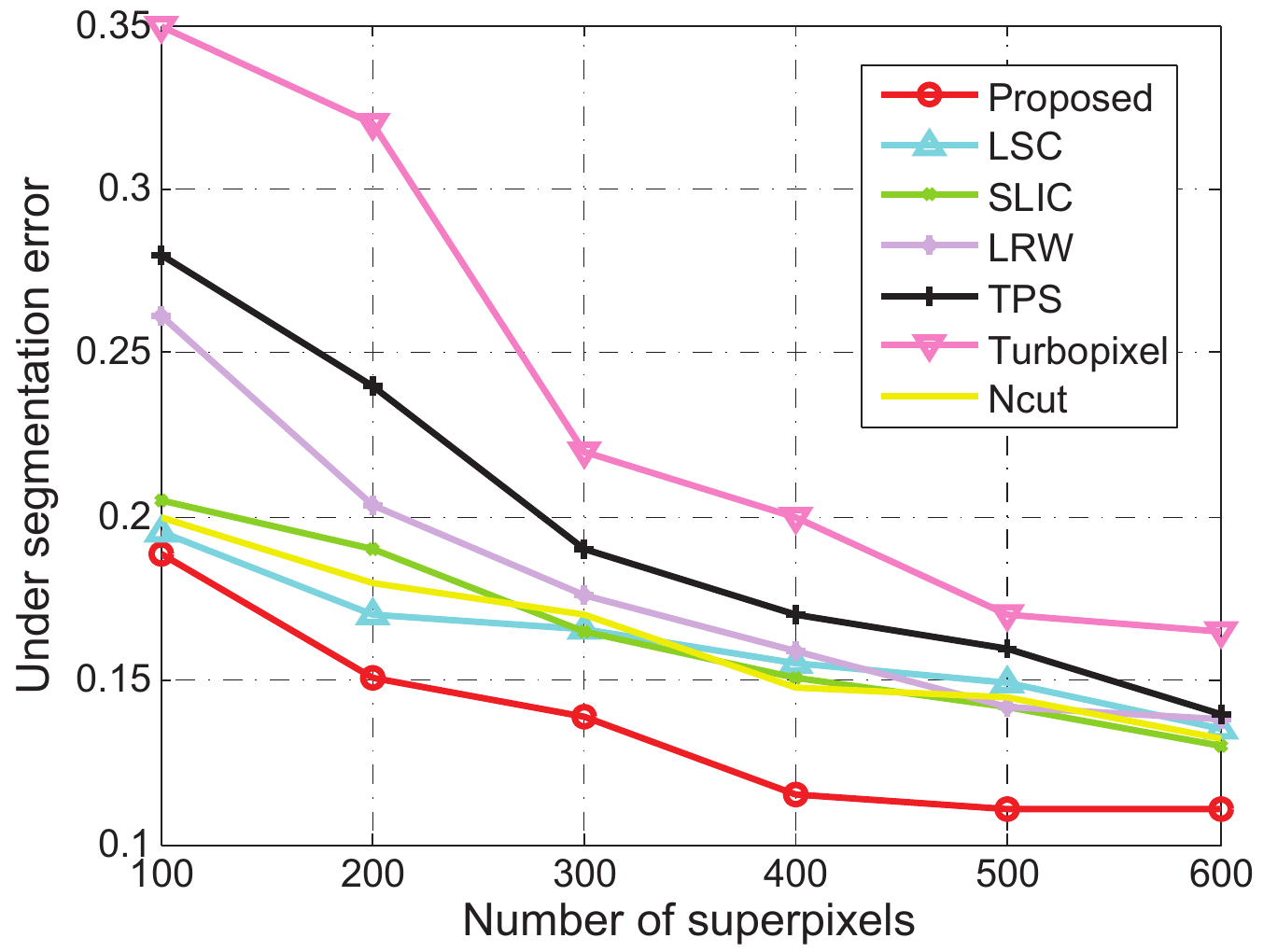}%
\label{USE-N}}
\hfil
\subfloat[BSDS test200]{\includegraphics[width=0.3\textwidth]{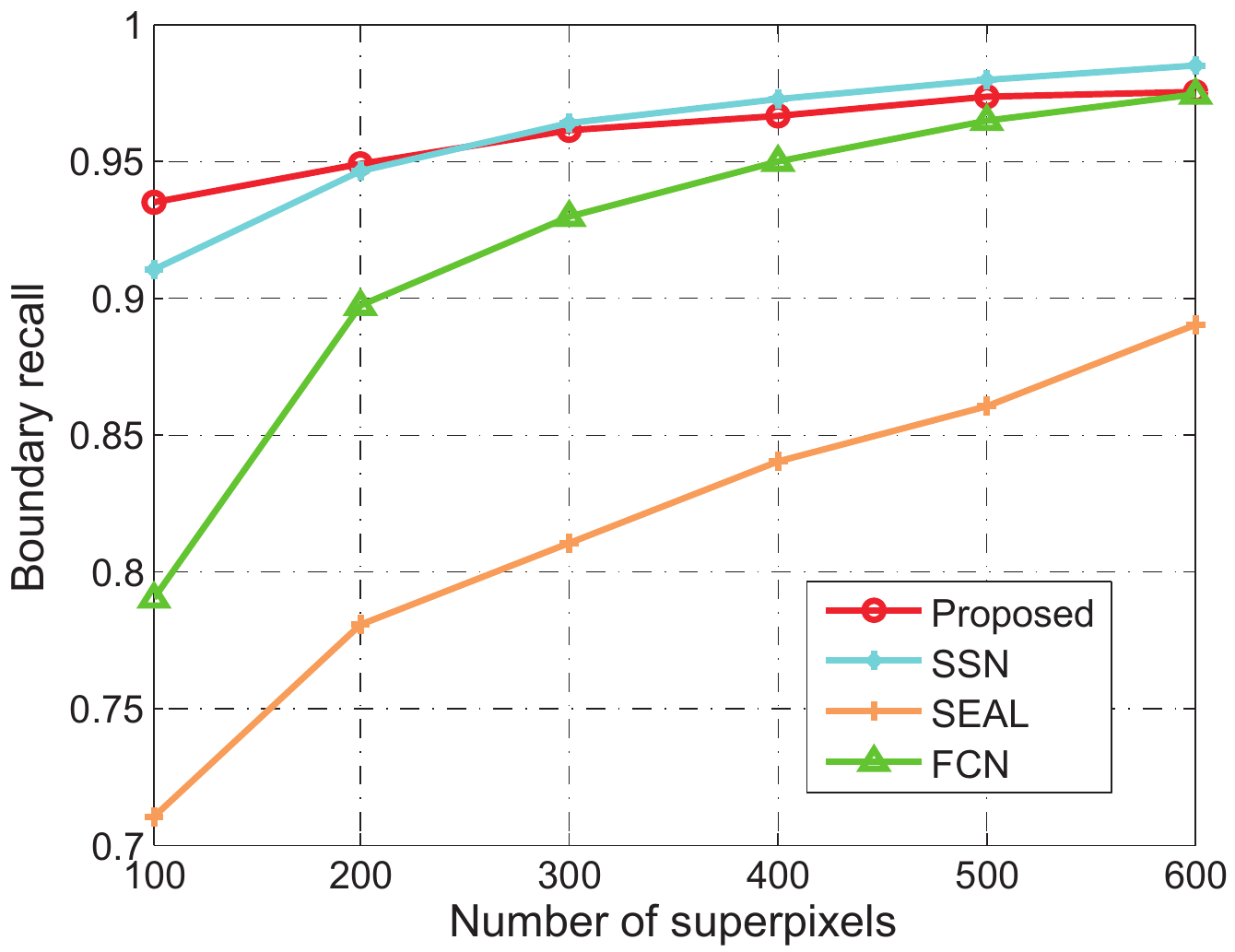}%
\label{ASA-N}}
\caption{Quantitative results. We compare our method with the state-of-the-art superpixel algorithms quantitatively by computing the BR, USE, and ASA scores on the BSDS500 in (a), (b), and (c). (d) and (e) are the ASA scores on DAVIS and Cityscapes datasets, respectively. (f) is the comparison of the BR vs. running time on BSDS500. (g) and (h) are the BR and USE score on NYUv2 datasets, respectively. (i) is the BR score compared with deep-learning based methods on BSDS test200.}
\label{fig-BSD}
\end{figure*}



\subsection{Parameter Selection}
The image size of BSDS500 dataset is $481 \times 321$. If every pixel is drawn as a sample, the dimension of the input is 154401. To speed up the computation, we use K-means to produce raw pixel units to keep the dimension of input data at thousand level. Let $r$ denote the ratio of the initial number of pixel units $n$ and the desired number of superpixels $K$. Fig. \ref{fig-ratio} shows the influence of $r$ with respect to different metrics. We can see that the segmentation quality keeps a relatively high level when $r$ is around 3. Indeed, blindly increasing the number of raw input units cannot improve the performance obviously, but increase the computation amount. Eventually we set the parameter $r$ to 3 in our experiments to trade off the computation speed and segmentation performance.
We have tried different ${\lambda _1}$ and ${\lambda _2}$, and find that excessively increasing the weight of ${\lambda _1}$ and ${\lambda _2}$ may reduce the performance.
Finally, ${\lambda _1}$ and ${\lambda _2}$ in (\ref{eq6}) are empirically set to $10^{-6}$ and $10^{-4}$, respectively (see more details in Section \ref{ablationsection}). The features are extracted from each pixel units, including color feature (average gray and R, G, B values of pixels unit), spatial feature (average coordinate of pixels unit), edge feature (average x and y components of gradient, gradient magnitude, and gradient direction) and texture feature (extracted by \cite{silva2015extended}). These features are concatenated and vectorized to ${\mathbf{X}}$.


\begin{figure*}[!t]
\centering
{\includegraphics[width=0.3\textwidth]{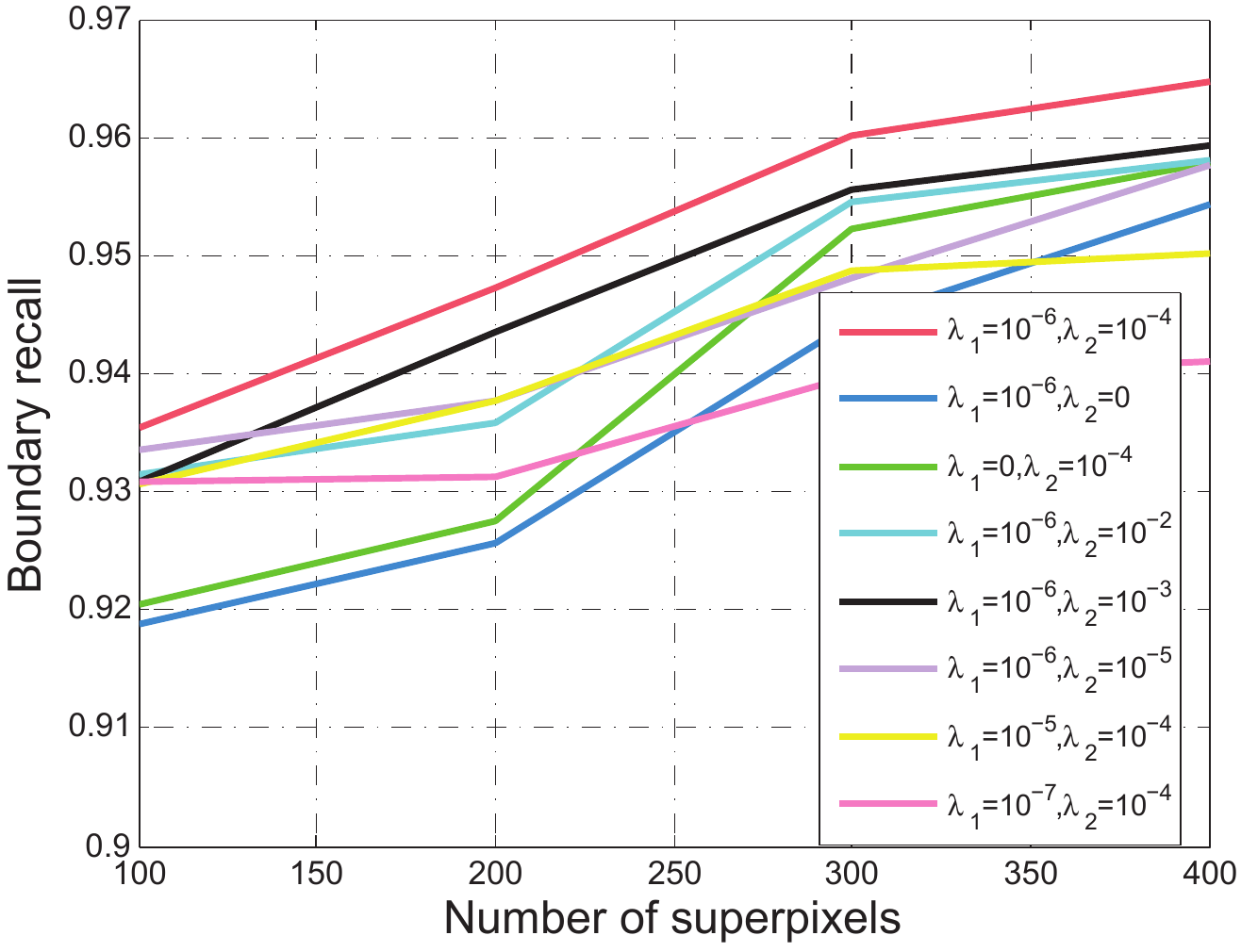}%
\label{ablation-BR}}
\hfil
{\includegraphics[width=0.3\textwidth]{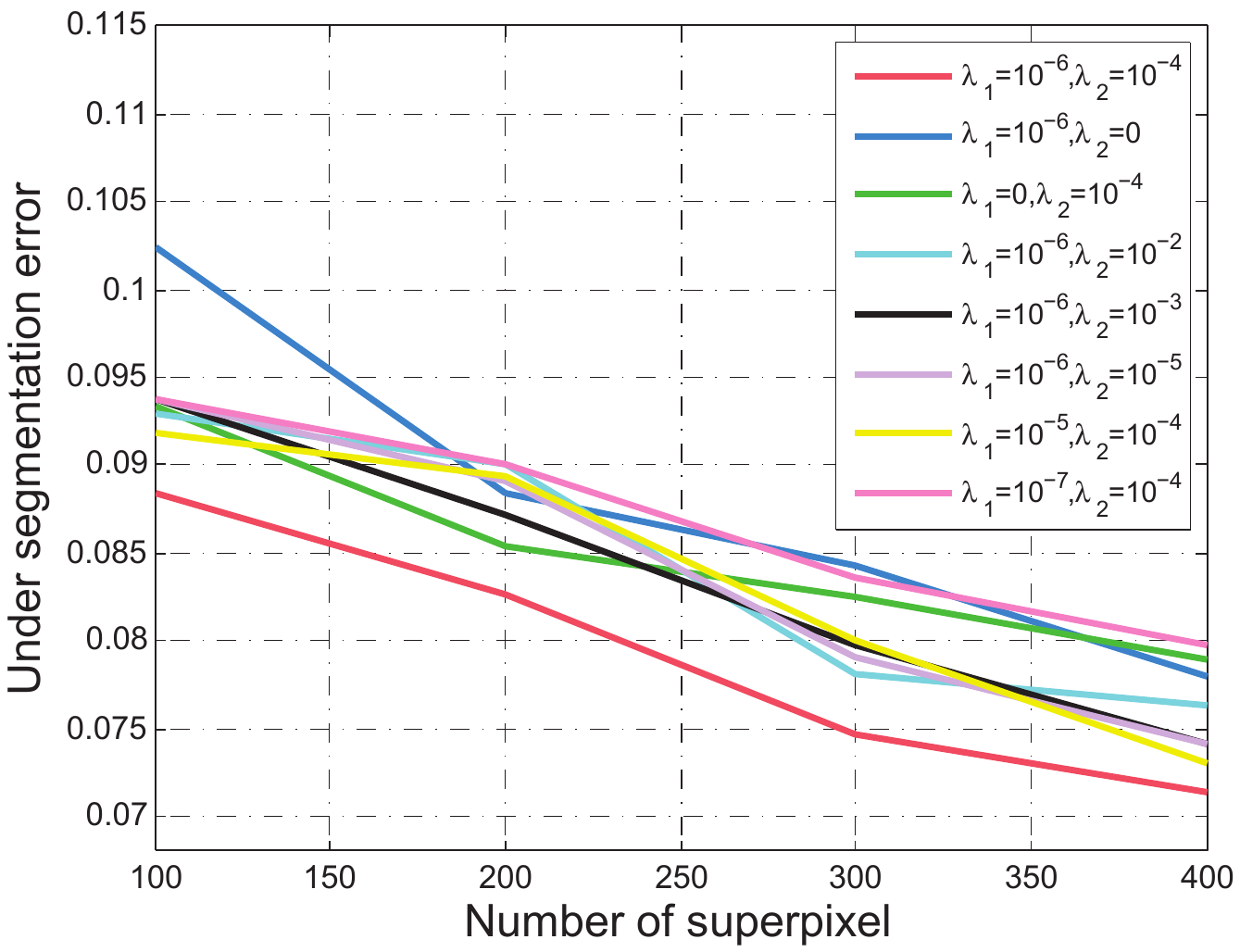}%
\label{ablation-USE}}
\hfil
{\includegraphics[width=0.3\textwidth]{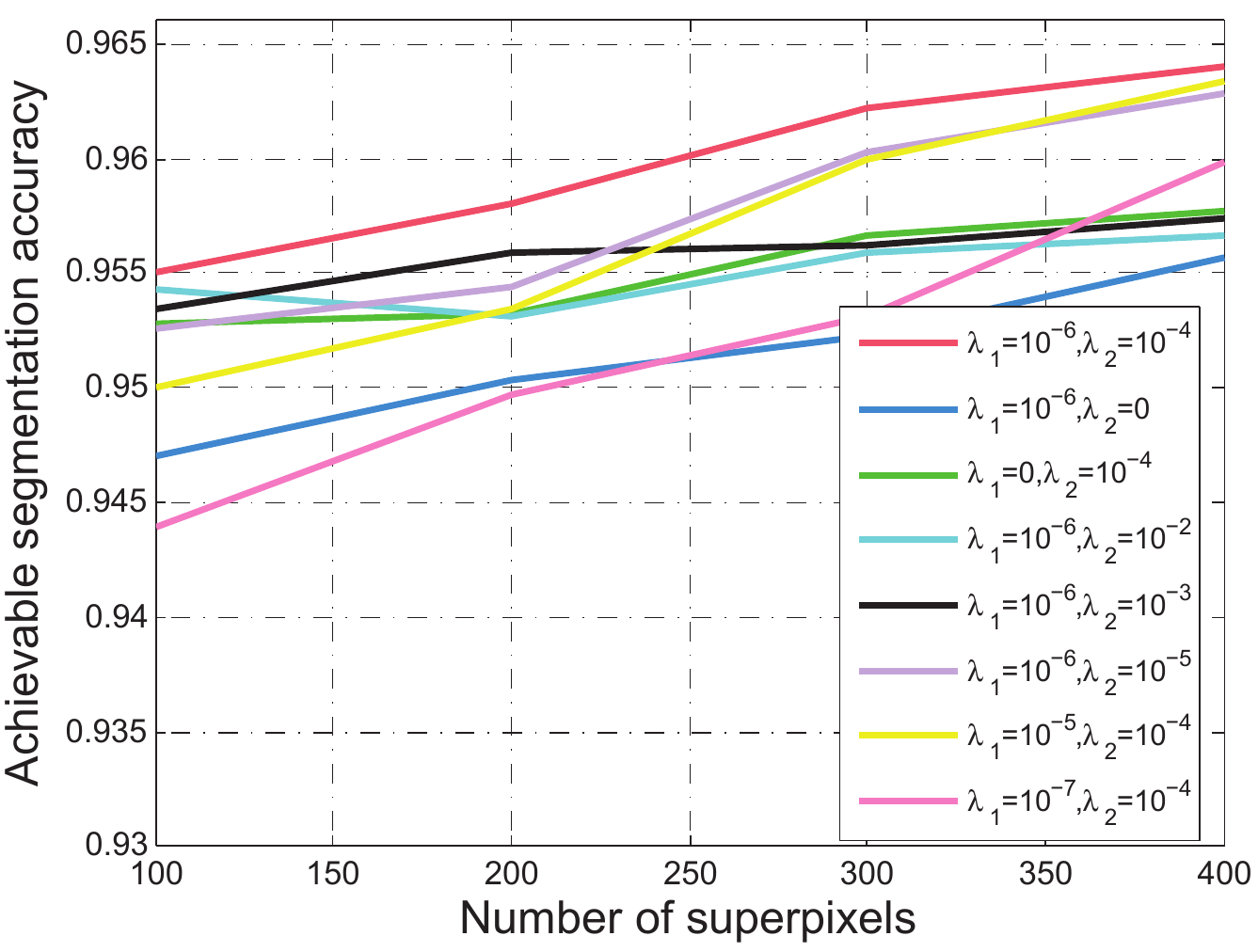}%
\label{ablation-ASA}}
\hfil

\caption{Ablation studies on BSDS500. When ${\lambda _1=10^{-6}}$, ${\lambda _2=10^{-4}}$, BR, USE, and ASA can achieve the highest score.}
\label{fig-ablation}
\end{figure*}

\subsection{Visual Comparison}
\label{result}
We compare our method with some state-of-the-art superpixel segmentation methods.
 Fig. \ref{unsupervised} illustrates some visual comparative results when the desired number of superpixels $K$ is set to 300. As can be seen, the proposed method maintains more detailed and complete object boundaries. For example, in the first image, our method has kept more detailed boundary information: most of the small spots in the butterfly wing are captured by our method, but almost unrecognizable by the compared methods.  For the second image, our method also catches the most complete and accurate boundaries, and moreover, the homogeneous regions of the zebra stripes have better coherence and clearer segmentation compared with others. In the third image, the boundaries of doggies are preserved more accurately than other methods. Ncut, LRW and TPS cannot catch most of the boundaries in the doggies' forehead and nose, while SLIC, FLIC and LSC can capture most boundaries but missing a little bit. Only the segments of our method can cover all the boundaries of foreheads, and the most boundaries of noses. The reason is that in our method, each superpixel is regarded as a subspace
 containing pixels with similar intrinsic attributes, such as similar texture, boundary, color and coherent spatial information.
  By this way, the pixels in the unit containing independent semantic information are more probably segmented into
  one superpixel, no matter how small the unit area is.


It is clear that other methods can also capture more image boundaries like ours if the desired superpixel number $K$ increases heavily.
In Fig. \ref{dot}, we investigate the minimum desired superpixel number $K$ for accurate spot segmentation in the butterfly image. The proposed method can segment the spot when $K=300$, while that of LRW, SLIC, and LSC are 1600, 1800, and 1600, respectively. This shows that our method can achieve more details and semantic information with less superpixels compared with other methods.

Subsequently, the proposed method is compared with three recent deep learning based methods, and Fig. \ref{SCS-deep} illustrates some visual results. It can be seen that our method can preserve more detailed boundaries than the deep-learning based methods. In the first image, only SEAL and our method segment the logo of the airplane. In the second image, both the two athletes' number ``1257'' and ``282'' are mostly recognizable in the proposed method, while in FCN, SSN, and SEAL, the number ``1257'' are unrecognizable. So our method is comparable with deep learning based methods, and even has superiority in detail preservation. It is also worth pointing out that our method is unsupervised, which does not need any ground truth information, while the compared deep-learning based methods all need some labeled samples that are costly to acquire.

\subsection{Quantitative Comparison}
\label{deep_compare}

Fig. \ref{fig-BSD} shows the quantitative comparisons with some state-of-the-art methods. Fig. \ref{fig-BSD} (a) and Fig. \ref{fig-BSD} (b) are the BR and USE metric results, which show that our method can largely improve the performance and outperforms all the other compared methods. Taking 300 superpixels for example, our method's minimum percentage gain (computed with the highest score of the compared methods) of USE is 62.5\%, while that of BR is 6.7\%. Fig. \ref{fig-BSD} (c), Fig. \ref{fig-BSD} (d), and Fig. \ref{fig-BSD} (e) are the ASA scores on BSDS500, DAVIS, Cityscapes datasets, respectively. In terms of ASA, the proposed method is competitive and improves the performance especially when the number of superpixels is small. Fig. \ref{fig-BSD} (g), and Fig. \ref{fig-BSD} (h) are the BR and USE score on NYUv2 dataset. The performance of our method also achieves the best performance in this dataset.
The minimum and maximum percentage gains of BR are 3.60\% and 41.15\%, respectively. The minimum and maximum percentage gains of USE are 3.43\% and 40.22\%, respectively. The proposed method can largely improve the segmentation quality.

Fig. \ref{fig-BSD} (i) shows the BR scores of our method and deep learning based methods FCN, SEAL and SSN on the 200 images of test in BSDS500. Compared with the deep learning based methods, the BR score of our method is higher than SEAL and FCN, and comparable with SSN. Moreover, our method performs better than SSN when the number of superpixels is less than 300.

Our method is implemented in Matlab and runs on a PC with a 3.6 GHZ Intel Core i7-6850K CPU and 16GB RAM. Fig. \ref{fig-BSD} (f) shows the trade-off between efficiency and performance compared with some state-of-the-art methods when desired superpixel number is 100 and 600. According to the BR scores, the proposed method can achieve the best segmentation performance, slightly slower than SLIC and LSC, but faster than others.

In conclusion, the quantitative results show that the proposed method performs better than the compared state-of-the-art methods by standard evaluation criteria.

\begin{table}[!t]
\caption{Effectiveness of the global and spatial regularization in the proposed model}
\label{table2}
\centering
\begin{tabular}{c|c|c|c|c}
\hline\hline
 $K$& Metric & $E$ + $GR$  & $E$ + $SR$ & $E$ + $GR$ + $SR$ \\
\hline
\multirow{3}*{100}&BR & 0.9188 & 0.9204 & \textbf{0.9355} \\
                ~ &USE & 0.1024 & 0.0934 & \textbf{0.0884} \\
                ~ &ASA & 0.9470 & 0.9528 & \textbf{0.9550} \\
\hline
\multirow{3}*{200}&BR & 0.9256 & 0.9276 & \textbf{0.9473} \\
                ~ &USE & 0.0884 & 0.0854 & \textbf{0.0827} \\
                ~ &ASA & 0.9504 & 0.9532 & \textbf{0.9580} \\
\hline
\multirow{3}*{300}&BR & 0.9442 & 0.9522 & \textbf{0.9602} \\
                ~ &USE & 0.0843 & 0.0805 & \textbf{0.0746} \\
                ~ &ASA & 0.9523 & 0.9567 & \textbf{0.9622} \\
\hline
\multirow{3}*{400}&BR & 0.9544 & 0.9578 & \textbf{0.9646} \\
                ~ &USE & 0.0789 & 0.0780 & \textbf{0.0714} \\
                ~ &ASA & 0.9557 & 0.9577 & \textbf{0.9640} \\
\hline\hline

\end{tabular}
\end{table}

\subsection{Ablation Study}
\label{ablationsection}
To verify the effectiveness of the global regularization ($GR$) and spatial regularization ($SR$) in the proposed model, we evaluate our method on BSDS500 by varying the trade-off parameters ${\lambda _1}$ (the weight of global term in (\ref{eq6})) and ${\lambda _2}$ (the weight of spatial term in (\ref{eq6})). Table \ref{table2} lists the results of our model without the $GR$ or $SR$ respectively with different K. $E$ is the error term in (\ref{eq6}).
It can be seen both the $SR$ and $GR$ are important to the proposed model, since $E + GR + SR$ always performs the best.
Moreover, $E + SR$ performs better than $E + GR$, indicating the importance of the spatial prior term.

Moreover, Fig. \ref{fig-ablation} illustrates the BR, USE, and ASA scores with different $\lambda_1$ and $\lambda_2$. From Fig. \ref{fig-ablation}, we can see that excessively increasing the weight of ${\lambda _2}$ reduces the performance of BR. When ${\lambda _1}=10^{-6}$ and ${\lambda _2}=10^{-4}$, all the metrics achieve the highest score. Therefore, we choose these two values for the parameters.
For DAVIS and Cityscapes datasets, we adopted the same parameter values and achieved excellent results.

\section{Conclusion}
\label{section6}
In this paper, we formulated image superpixel segmentation as a subspace clustering problem based on the assumption that an input image is constituted of several subspaces, and pixels with similar intrinsic attributes lie in the same subspace. We designed a novel local regularization term to enforce the spatial correlation, and the proposed spatially constrained subspace clustering model is able to generate content-aware superpixels with more detailed boundaries. The proposed method has been compared with some state-of-the-art methods, and the experiments on different standard datasets demonstrate that the proposed method achieves superior performance both quantitatively and qualitatively.

\ifCLASSOPTIONcaptionsoff
  \newpage
\fi



\bibliographystyle{IEEEtran}
\bibliography{IEEEabrv,ref}
\end{document}